\documentclass[11pt,letterpaper,english]{article}
\usepackage{microtype}
\usepackage{float}
\usepackage{amsmath} 
\usepackage{amsthm}
\usepackage{mathrsfs}
\usepackage{graphicx} 
\usepackage{caption}
\usepackage{setspace}
\usepackage{amssymb}
\usepackage{multirow}
\usepackage{rotating}
\usepackage{array}
\usepackage{booktabs}
\usepackage{multibib}
\usepackage{bibunits}
\usepackage{xr}
\usepackage{algorithm}
\usepackage{algpseudocode}
\usepackage{color}
\usepackage[small,compact]{titlesec} 
\DeclareMathAlphabet{\mathcalligra}{T1}{calligra}{m}{n}
\usepackage[sectionbib,round]{natbib}
\usepackage{mdframed}
\usepackage{tcolorbox}

\usepackage{times}
\usepackage{abstract}

\usepackage[hang,flushmargin]{footmisc}
\usepackage{fullpage} 
\usepackage{hyperref}
\usepackage[compact]{titlesec}
\usepackage[section]{placeins}
\usepackage{footnote}
\usepackage{epstopdf}
\usepackage{placeins}
\usepackage[toc,page]{appendix}
\usepackage{graphicx}
\usepackage{subcaption}
\usepackage{amsmath} 
\usepackage{amsthm}
\usepackage{graphicx} 
\usepackage{blkarray}
\usepackage{stackengine}
\usepackage{accents}
\usepackage{enumitem}
\usepackage[mathscr]{euscript}

\defaultbibliographystyle{abbrvnat} 
\defaultbibliographystyle{plainnat} 
\defaultbibliography{ref} 
\usepackage{tabularx}

\hypersetup{
    colorlinks=true,      
    linkcolor=MyDarkBlue,      
    citecolor=MyDarkBlue,      
    filecolor=MyDarkBlue,      
    urlcolor=MyDarkBlue        
}
\definecolor{MyDarkBlue}{RGB}{158,0,0}

\makeatletter
\def\UrlAlphabet{%
      \do\a\do\b\do\c\do\d\do\e\do\f\do\g\do\h\do\i\do\j%
      \do\k\do\l\do\m\do\n\do\o\do\p\do\q\do\r\do\s\do\t%
      \do\u\do\v\do\w\do\x\do\y\do\z\do\A\do\B\do\C\do\D%
      \do\E\do\F\do\G\do\H\do\I\do\J\do\K\do\L\do\M\do\N%
      \do\O\do\P\do\Q\do\R\do\S\do\T\do\U\do\V\do\W\do\X%
      \do\Y\do\Z}
\def\UrlDigits{\do\1\do\2\do\3\do\4\do\5\do\6\do\7\do\8\do\9\do\0}
\g@addto@macro{\UrlBreaks}{\UrlOrds}
\g@addto@macro{\UrlBreaks}{\UrlAlphabet}
\g@addto@macro{\UrlBreaks}{\UrlDigits}
\makeatother

\theoremstyle{definition}

\theoremstyle{definition}

\usepackage[utf8]{inputenc}
\usepackage{longtable}
\usepackage{graphicx}
\usepackage{epstopdf}
\usepackage{bbm}
\usepackage{dsfont}
\usepackage{comment}
\usepackage{pgfplots}
\usepackage{xfrac}
\usepackage{algorithm}
\usepackage{algpseudocode}
\usepackage{siunitx}

\usepackage{pifont}

\usepackage{caption}
\def\bE{\mathbb{E}}

\theoremstyle{plain}

\newcommand{\squishlist}{
   \begin{list}{$\bullet$}
    { \setlength{\itemsep}{0pt} \setlength{\parsep}{1pt}
      \setlength{\topsep}{1pt} \setlength{\partopsep}{1pt}
      \setlength{\leftmargin}{1.5em} \setlength{\labelwidth}{1em}
      \setlength{\labelsep}{0.5em} } }

\newcommand{\squishlisttwo}{
   \begin{list}{$\bullet$}
    { \setlength{\itemsep}{0pt} \setlength{\parsep}{0pt}
      \setlength{\topsep}{0pt} \setlength{\partopsep}{0pt}
      \setlength{\leftmargin}{1em} \setlength{\labelwidth}{1.5em}
      \setlength{\labelsep}{0.5em} } }

\newcommand{\squishend}{
    \end{list}  }

\title{LOLA: LLM-Assisted Online Learning Algorithm \\for Content Experiments}

\author{Zikun Ye \thanks{We would like to thank Yuting Zhu and the participants of the QME 2024, UW--UBC Conference 2024, Informs Marketing Science 2024, and China India Insights Conference 2024 for feedback. Thanks are also due to the marketing seminar attendees at the University of Southern California for their comments that have significantly improved the paper. Please address all correspondence to: zikunye@uw.edu and hemay@uw.edu.} \\ \textit{University of Washington} \and Hema Yoganarasimhan \\ \textit{University of Washington} \and Yufeng Zheng \\ \textit{University of Toronto}}

\pgfplotsset{compat=1.18} 

\begin{document}

\maketitle
\begin{abstract}
\begin{singlespace}

Modern media firms require automated and efficient methods to identify content that is most engaging and appealing to users. Leveraging a large-scale dataset from Upworthy (a news publisher), which includes 17,681 headline A/B tests, we first investigate the ability of three pure-LLM approaches to identify the catchiest headline: prompt-based methods, embedding-based methods, and fine-tuned open-source LLMs. Prompt-based approaches perform poorly, while both OpenAI-embedding-based models and the fine-tuned Llama-3-8B achieve marginally higher accuracy than random predictions. In sum, none of the pure-LLM-based methods can predict the best-performing headline with high accuracy. We then introduce the LLM-Assisted Online Learning Algorithm (LOLA), a novel framework that integrates Large Language Models (LLMs) with adaptive experimentation to optimize content delivery. LOLA combines the best pure-LLM approach with the Upper Confidence Bound algorithm to allocate traffic and maximize clicks adaptively. Our numerical experiments on Upworthy data show that LOLA outperforms the standard A/B test method (the current status quo at Upworthy), pure bandit algorithms, and pure-LLM approaches, particularly in scenarios with limited experimental traffic. Our approach is scalable and applicable to content experiments across various settings where firms seek to optimize user engagement, including digital advertising and social media recommendations. 

\end{singlespace}
\end{abstract}


\newpage
\begin{bibunit}

\section{Introduction}
\label{sec:intro}

Digital content consumption has seen unprecedented growth, leading to a proliferation of content across various platforms. In today’s real-time digital environment, media firms and news publishers need automated and efficient methods to determine which content generates high user engagement and platform growth. This includes identifying the most appealing articles, the most attractive headlines, and the catchiest cover images \citep{NYTHeadline2017, NYTimes2019}. Traditionally, media firms and publishers have relied on experimentation-based approaches to address this problem. Broadly speaking, there are two types of experimentation styles---(1) Standard A/B tests and (2) Online learning algorithms or bandits.\footnote{We use the terms adaptive experimentation, online learning algorithm, and bandits interchangeably throughout the paper.}

A/B test is the most straightforward experimentation method, where a firm allocates a fixed portion of traffic to different treatment arms, assesses the results, and then goes with the best-performing arm. This is also known as the Explore and Commit strategy, or E\&C, where the firm explores for a fixed period using A/B tests and then commits to one treatment based on the results of those tests. This approach is widely used in the publishing industry. For instance, The New York Times recently built a centralized internal A/B test platform, ABRA (A/B Reporting and Allocation architecture), allowing different teams to experiment with their content; see \cite{NYTimes2024} for more details. Another example is Upworthy, which has implemented extensive A/B tests to choose headlines for news articles \citep{matias2021upworthy}. The main advantage of this approach is that it is trustworthy---with a sufficiently large amount of traffic allocated to the A/B test, the resulting inference is unbiased and accurate. However, a major drawback of this approach is the wastage of traffic---by assigning equal traffic to all the treatment arms during the exploration phase (including the poorly performing ones), the firm incurs high regret or low profits. This is especially problematic in the news industry since news articles tend to have short lifetimes and become stale quickly (typically within a day or two). Therefore, if a firm wastes a lot of traffic to learn the best article/headline, the article itself might become irrelevant by the end of the A/B test.

The second approach, online learning algorithms, or bandits, are able to address some of these problems. These algorithms dynamically adjust traffic allocation to articles/headlines based on their features, past performance, and associated uncertainty. This method provides a more efficient way to optimize content delivery by continually learning and adapting. For instance, both Yahoo News and The New York Times have successfully implemented contextual bandit algorithms to identify content that maximizes user engagement \citep{li2010contextual, NYTimes2019}.  Online learning algorithms typically outperform A/B tests in terms of overall regret or reward. These algorithms reduce the extent of exploration over time by moving traffic away from poorly performing treatments dynamically. As a result, they tend to switch to the best-performing arms quickly and incur lower regret compared to standard E\&C strategies. This has been established both theoretically \citep{lattimore2020bandit} as well empirically \citep{li2010contextual}.\footnote{Similarly, in the context of digital advertising, \cite{schwartz2017customer} use field experiments to demonstrate that adaptive algorithms outperform the standard balanced A/B test.} 





Nevertheless, online algorithms also have notable drawbacks. In standard content experiments, these algorithms suffer from a cold-start problem. That is, firms typically start these adaptive experiments assuming that all arms are equally likely to be the best-performing ones. This can lead to a situation where more-than-necessary traffic is allocated to sub-optimal arms, leading to higher regret. Additionally, due to their probabilistic nature, these algorithms can inadvertently converge on a sub-optimal arm, especially when the experimental traffic is small. These issues arise because online algorithms are designed for large traffic regimes, where the initialization of arms has little impact on the asymptotic regret. However, as discussed earlier, there is limited traffic available for experimentation in the media industry since content becomes stale quickly. Therefore, the negative effects of the ``equally effective'' initialization do not diminish sufficiently during the time horizon of interest, leading to higher regret.


Given these issues with content experimentation, the main question that this paper asks and answers is whether we can leverage the recently developed Large Language Models (LLMs) to enhance current content experimentation practices and simplify the process of identifying appealing content. LLMs, trained on extensive human-generated data, have demonstrated the ability to mimic human preferences and behavior in a variety of consumer research tasks \citep{li2024frontiers, brand2023using}. Specifically, in the context of news articles, \cite{yoganarasimhan2024feeds} show that LLMs can correctly predict the polarization of news content. Hence, a relevant question is whether firms can directly use LLMs to predict the appeal of different content, potentially replacing traditional experimentation. This approach could significantly reduce the costs associated with experimentation while simplifying the content delivery processes.


To address this question, we employ a large dataset of content experiments by the publishing firm Upworthy conducted in the 2013--2015 timeframe. The dataset consists of 17,618 headline tests spanning 77,245 headlines, 277 million impressions, and 3.7 million clicks. The goal of each A/B test was to identify the best headline among a set of candidate headlines (for a given article). This dataset is an excellent test-bed for our study for a few reasons -- (1) it is based on an extremely large number of A/B tests over a high volume of impressions, (2) each A/B test received a relatively large number of impressions, allowing for accurate estimates of each headline's performance, and (3) it is sourced from a real media firm and represents the actions of a very large number of readers over a sufficiently long period to offer meaningful insights.

In the first part of the paper, we use this dataset to examine the extent to which pure-LLM-based approaches can accurately predict headline attractiveness. For this exercise, we test the different approaches' accuracies, defined as the success rate of identifying the headline with the highest click-through rate (CTR) in a given A/B test.


We consider the three most widely used LLM-based approaches for this analysis---(1) Prompt-based approaches, (2) LLM text-embedding approaches, and (3) Fine-tuning approaches. For prompt-based methods, we consider two approaches: (a) Zero-shot prompting and (b) In-context learning. Zero-shot prompting is a technique in which an LLM generates responses without being explicitly trained on specific examples. This is the most common way in which most users interact with the LLM. In-context learning involves providing the LLM with a few demonstrations of similar tasks before generating responses for a specific task \citep{min2022rethinking, xie2021explanation}. We find that GPT-3.5 performs as poorly as a random guess for both prompt-based approaches. GPT-4 performs marginally better---it achieves an accuracy of 37.85\% for zero-shot prompting and 38-40\% for in-context learning. Overall, this suggests that prompt-based approaches cannot aid/replace content experiments in any meaningful way.

Next, we turn to the second approach: text embeddings. Specifically, we transform headlines into embedding vectors using the most recent embedding model from OpenAI. We then use the embedding as the input, along with the CTR as the output, to train CTR prediction models, including linear regression and multilayer perceptron (a type of neural network). After getting predicted CTRs, we choose the headline with the highest predicted CTR as the winner. We find that OpenAI's embeddings combined with simple linear regression can achieve around 46.28\% accuracy, while more complex neural networks do not improve performance over the simple linear model. Finally, we fine-tune a state-of-the-art open-source LLM --- Llama-3 with 8 billion parameters, using low-rank adaptation (LoRA) \citep{hu2021lora}. This approach performs the best, with 46.86\% accuracy, which is a small improvement over the embedding-based approach. 
However, even the best-fine-tuned LLMs are unable to perfectly predict which types of content are most appealing to users. As such, none of the LLM-based approaches can match the accuracy of experimentation-based approaches.

Therefore, in the second part of the paper, we propose and evaluate a novel framework for content experiments that combines the strengths of LLM-based approaches with the advantages of online learning algorithms, termed LOLA (\textbf{L}LM-Assisted \textbf{O}nline \textbf{L}earning \textbf{A}lgorithm). Our main insight is to use predictions from an LLM model as priors in online algorithms to optimize experimentation. While the general idea that priors can impact regret and rewards has existed for some time \citep{bubeck2013prior, russo2014learning}, these papers typically assume that the quality and distribution of priors are known. However, a key challenge in our setting is that the LLM predictions cannot be naively treated as priors since we do not know their prediction error or theoretical properties. Therefore, we build on the recently proposed 2-UCBs algorithm by \citet{gur2022adaptive} to overcome this challenge. The original 2-UCBs algorithm was designed for a setting where the experimenter has access to auxiliary data and knows both the size of this auxiliary sample ($n^{aux}$) and knowledge of the prediction error in this sample. Our innovation is to extend this algorithm to accommodate LLM predictions by treating these predictions {\it as if} they came from a ``pseudo-sample" and then to fine-tune the size of this pseudo-sample for a given application based on prior data using hyper-parameter tuning. Thus, we are able to combine LLM predictions with online experimentation without making strong assumptions on the theoretical properties of the LLM-based predictors.

LOLA involves two steps. First, we train an LLM-based CTR prediction model using LoRA fine-tuning. The second step integrates these predictions into online learning algorithms. LOLA is a general framework, and depending on the specific goals of the experimentation, it can be adapted for different prediction models in the first step and different online algorithms in the second step. For example, if the goal is to minimize regret (i.e., maximize total reward/clicks), we propose a modified Upper Confidence Bound algorithm called LLM-Assisted 2-Upper Confidence Bounds (LLM-2UCBs) that is detailed in $\S$\ref{ssec:lola}. The idea of LLM-2UCBs is intuitive: we can view LLM-based CTR predictions as auxiliary samples before the start of the online algorithm (e.g., a 1\% CTR can be viewed as 100 impressions with 1 click or 1,000 impressions with 10 clicks). After fine-tuning this auxiliary impression size hyperparameter, we incorporate these auxiliary samples into the standard UCB algorithm \citep{auer2002finite} to obtain the reward estimator and shrink the upper confidence bound. 

LOLA combines the advantages of both experimentation and LLMs, making it accurate and efficient. By integrating an LLM-based model to predict CTRs before the online deployment phase, we avoid wasting impressions on poorly performing headlines, especially in the initial stage when we face a cold-start problem. On the other hand, through experimentation, we can correct the errors in LLM predictions. This hybrid approach ensures that while LLMs provide reasonable initial predictions, ongoing experimentation refines and improves overall performance, making LOLA an effective solution for optimizing content delivery. 

We present a comprehensive evaluation of LOLA and compare its performance with three natural benchmarks using the Upworthy dataset: (1) Explore and Commit, (2) a pure online learning algorithm, and (3) a pure LLM-based model. The first benchmark, E\&C, was the status quo at Upworthy during the time of data collection, wherein the editors first ran an A/B test on a set of headlines for a fixed set of impressions and then displayed the winner for the rest of the traffic. For the pure online learning algorithm, we use the standard Upper Confidence Bound (UCB) algorithm, which is initialized with uniform CTRs. For the pure LLM-based approach, we use the LoRA fine-tuned Llama-3 to predict the CTRs of headlines and select the one with the highest CTR for all impressions, bypassing the experimentation phase. In LOLA, we use the same CTR prediction model as that used in the pure LLM approach and then employ LLM-2UCBs to maximize accumulated clicks. We find that LOLA outperforms all the baseline approaches, though the next-best algorithm varies based on the length of the time horizon (keeping the number of headlines fixed). 

Specifically, when time horizons are small, both LOLA and the pure LLM-based approach beat the pure experimentation-based approaches (i.e., E\&C and the UCB algorithm) due to relatively accurate LLM-based CTR predictions, and lack of sufficient time for the experimentation to provide meaningful updates. Specifically, both LOLA and the pure LLM outperform E\&C by 8-9\% and UCB by 6-7\% in this scenario. However, as the number of impressions/time horizon grows, the performance of the pure LLM-based approach falls behind other experimentation-based methods because pure LLM-based approaches are static and do not take advantage of experimentation. In this case, LOLA beats pure LLM methods and E\&C by 4--5\%, and UCB by 2-3\%. In sum, our experiments demonstrate that LOLA is able to leverage the power of LLMs early in the horizon and then build on it to take advantage of experimentation over time, thereby bringing together the strengths of both methods. 

Finally, while we focus on regret minimization in a stochastic bandit setting given the Upworthy context, LOLA is a general-purpose framework and can easily accommodate different goals and/or settings. We provide variants of the standard LOLA (and results on the empirical performance of these variants when applicable) for a few natural extensions, e.g., for Bayesian bandit settings where the second stage algorithm can be Thompson Sampling instead of UCB, for the best arm identification problem where the goal is to identify the best content/headline rather than regret minimization, and for settings where the firm has access to user/contextual features that can be used to personalize content delivery.

In summary, our paper makes three key contributions to the literature. First, from an empirical perspective, the paper provides evidence that LLMs (or, more precisely, the current vintage of LLMs) cannot replace experimental approaches in the media and digital marketing industry. Second, from a methodological perspective, we develop a novel framework, LOLA, to leverage LLMs for improving content experimentation in digital settings. To the best of our knowledge, this is the first paper that proposes combining LLMs with adaptive experimentation techniques. We demonstrate the value of our approach using a large-scale dataset from Upworthy when compared to baseline methods. Our work thus provides a foundation for future research into more advanced algorithms and offers insights for real-world validation in industry settings. Third, from a managerial perspective, LOLA can be used in a broad range of settings where firms need to decide which content to show. While we focus on the news/publishing industry, given our empirical context, the LOLA framework is general and can easily extend to other settings, including digital advertising, email marketing, and website design. Further, the method is cost-effective, can be built on open-source LLMs (alleviating data privacy concerns), and is easily deployable across different domains once fine-tuned on relevant datasets. 

\section{Upworthy Data and Experiments}
\label{sec:upworthy_data}


\begin{table}[t]
\centering
\begin{tabularx}{\textwidth}{cXcc}
\toprule
Test ID&  Headline&  Impressions& Clicks\\
\midrule
1& New York's Last Chance To Preserve Its Water Supply  & 2,675 & 15 \\
1& How YOU Can Help New York Stay Un-Fracked In Under 5 Minutes & 2,639 & 19 \\
1& Why Yoko Ono Is The Only Thing Standing Between New York And Catastrophic Gas Fracking &  2,734 & 34\\
2&  If You Know Anyone Who Is Afraid Of Gay People, Here's A Cartoon That Will Ease Them Back To Reality & 4,155 & 120\\
2& Hey Dude. If You Have An Older Brother, There's A Bigger Chance You're Gay & 4,080 & 41\\
 \bottomrule
\end{tabularx}
\caption{Samples of Upworthy experiments' results for headlines. We display the key columns for our analysis, including the ID for the A/B tests, the text of headlines, and the number of impressions and clicks in the test.}
\label{tab:data in upworthy}
\end{table}

We now discuss the data sourced from Upworthy, a U.S. media publisher known for its extensive use of A/B tests in digital publishing. Upworthy conducted randomized experiments with each article published, exploring different combinations of headlines and images to determine which elements most effectively led to higher clicks. The archival record from January 24, 2013, to April 30, 2015, detailed in \cite{matias2021upworthy}, demonstrates how the packaging of headlines and images played a crucial role in Upworthy's growth strategy. Upworthy's editorial team created several versions of headlines and/or images for each article (internally called ``package'' by Upworthy). A package is defined as one treatment or arm for an article and consists of a headline, image, or a combination of both. Editors would first choose several of what they believed were the most promising packages to be tested. Then, they A/B tested these packages or treatments to identify which one resonated the best with their audience. During the A/B test, users only saw the headline (and an accompanying image in some cases), but not the article itself. Upon clicking on the headline, they were taken to the article. As such, the content of the article itself did not have any effect on users' clicking behavior. Upworthy's A/B test system recorded how many impressions and clicks each package received. Table \ref{tab:data in upworthy} shows two examples of A/B tests with the relevant columns. The full details of all the A/B tests and their results are available from the \href{https://osf.io/jd64p/}{Upworthy Research Archive}. 


We now describe the details of this dataset and discuss our pre-processing procedure. In the original dataset, there are 150,817 tested packages from 32,487 deployed A/B tests. The dataset records a total of 538,272,878 impressions and 8,182,674 clicks. All these statistics are also listed in the top panel of Table \ref{tab:upworthy_stats}. These summary statistics indicate that during this period, each A/B test had an average of 4.64 packages, with each package receiving an average of 3,569 impressions and 54.26 clicks, resulting in an average click-through rate (CTR) of 1.52\%. Within an A/B test, each package/treatment arm had the same probability of receiving an impression; so the number of impressions received by all the packages within a test is approximately the same. However, the actual number of impressions varies across tests, with the first quartile at 2,745, the median at 3,117, and the third quartile at 4,089. This is due to the relatively straightforward implementation of A/B tests, i.e., Upworthy did not conduct any power analysis in advance to determine the traffic for a given test, as confirmed by \cite{matias2021upworthy}.

\begin{table}[ht]
\centering
\begin{tabular}{l|l|c}  
\toprule
\multirow{4}{*}{Original data} & \# of tests & 32,487 \\
 & \# of packages & 150,817 \\
 & \# of impression in tests & 538,272,878 \\
 & \# of clicks in tests & 8,182,674 \\
 \midrule
\multirow{7}{*}{Headline tests data} & \# of tests & 17,681 \\
 & \# of packages & 77,245 \\
 & \# of impression in tests & 277,338,713 \\
 & \# of clicks in tests & 3,741,517 \\
 \midrule
\multirow{1}{*}{} & \# of impressions (both test and non-test impressions) & 2,351,171,402 \\
\bottomrule
\end{tabular}
\caption{Basic summary statistics of Upworthy data during the time window between January 24, 2013 and April 30, 2015}
\label{tab:upworthy_stats}
\end{table}

Among these tests, some were conducted to test different images. However, the dataset does not allow us to trace back the actual images used in the tests. Therefore, we focus on the headline tests and filter out all the image tests. After this filtering, there are 17,681 tests, totaling 77,245 packages. On average, each article was tested with 4.37 headlines. Note that after filtering, each package or treatment simply refers to a unique headline; and therefore, we use the phrases package and headline interchangeably in the rest of the paper. The distribution of the number of packages/headlines for each article is shown in Table \ref{tab:dist_package}. After filtering, the dataset includes 277,338,713 impressions and 3,741,517 clicks; see the middle panel of Table \ref{tab:upworthy_stats}. We also report the total number of impressions for the entire Upworthy website, as recorded in their data from Google Analytics. We see that the impressions attributable to the tests constitute only 22.9\% of the total traffic.

\begin{table}[ht]
\centering
\begin{tabular}{@{}c|cc@{}}
\toprule
\# of headlines in one test & \# of tests & \% of samples \\
\midrule
2 & 1,619 & 9.16 \\
3 & 939 & 5.31 \\
4 & 8,836 & 49.97 \\
5 & 2,964 & 16.76 \\
6 & 2,685 & 15.19 \\
7 or more & 638 & 3.61 \\
\bottomrule
\end{tabular}
\caption{Distribution of number of headlines tested, across A/B tests.}
\label{tab:dist_package}
\end{table}

To illustrate that learning which headline is best is a non-trivial task, we conduct a survey where we give human users pairs of headlines from the set of headline pairs that are significantly different from each other and ask them to identify the catchier headline. Even within this set of headline pairs, we find that the respondents' accuracy in the survey was not significantly better than random guessing. This further highlights the fact that the task of identifying engaging content is inherently challenging, even for humans. See Web Appendix $\S$\ref{appsec:survey} for more details of the survey and its results.\footnote{Note that just because an average reader cannot discern/identify the most engaging or catchiest headline among a pair/set of headlines, it does not follow that such differences do not exist. Indeed, the whole premise of the experimentation culture in digital firms relies on the idea that while managers/lay users cannot correctly predict which version of an ad creative/promotion/website layout will be more engaging, such differences exist and that identifying these differences and implementing the best-performing arm is important and will lead to better consumer engagement and business outcomes.}

\section{Pure-LLM-Based Methods}
\label{sec:pure_llm}

We divided the 17,681 headline tests into three folds: 70\% for training, 10\% for hyperparameter fine-tuning of the algorithms (as discussed in $\S$\ref{sec:LOLA}), and 20\% for testing. This sample split is used consistently throughout the remainder of the paper unless stated otherwise. Specifically, in this section, we have 12,376 headline tests as a training set and 3,263 headline tests as a test set.\footnote{Note this test set is slightly smaller than 20\% of headline tests. This is because, to avoid information leakage, we deleted the duplicated tests that had the same headlines in the training dataset. This happens because Upworthy allowed additional experiments for the same headline under a new experiment ID, and the editor may require an extra test to make an informed decision.} 

In the rest of this section, we investigate the performance of three widely used LLM-based approaches---(1) GPT prompt-based approaches, including zero-shot prompting and in-context learning, (2) embedding-based models, and (3) fine-tuning of LLMs.


\subsection{Prompt-based Approaches}
\label{sec:prompts_incontext}

Prompt-based approaches use GPT prompts to select which headline is the catchiest, and there is no modeling or explicit fine-tuning involved. Prompt-based approaches have been the standard use-case for AI models in the marketing literature so far and have been used to examine if and when LLMs can simulate consumer behavior in market research studies \citep{gui2023challenge, brand2023using, li2024frontiers}. We consider two prompt-based approaches: (1) Zero-shot prompting and (2) In-context learning. 

\subsubsection{Zero-shot Prompting}
\label{sssec:zero-shot}
Zero-shot prompting is a technique in which an LLM generates responses or performs tasks without being explicitly provided by any specific examples. We evaluate the performance using zero-shot prompting based on OpenAI's GPT models. We do this through OpenAI's API in Python, which offers various parameters to customize and control the model's behavior and responses. The key parameters include specifying which GPT model to use (such as GPT-3 or GPT-4) and the system parameters, which guide the type of responses generated. The user's input or query is contained in the user parameter, while the assistant parameter holds the model's previous responses to maintain context in an ongoing conversation. Additionally, we set the temperature parameter to be 0, which indicates there is no randomness in the GPT's response for reproducibility.\footnote{The temperature parameter scales and controls the randomness of the model's responses. A larger value results in a more diverse and creative output, while a lower temperature makes the output more deterministic and focused. The default setting with temperature is 1.0. We also tested the default temperature, and we did not observe any significant difference in accuracy.}



\begin{figure}[ht]
    \centering
    \includegraphics[width=0.8\linewidth]{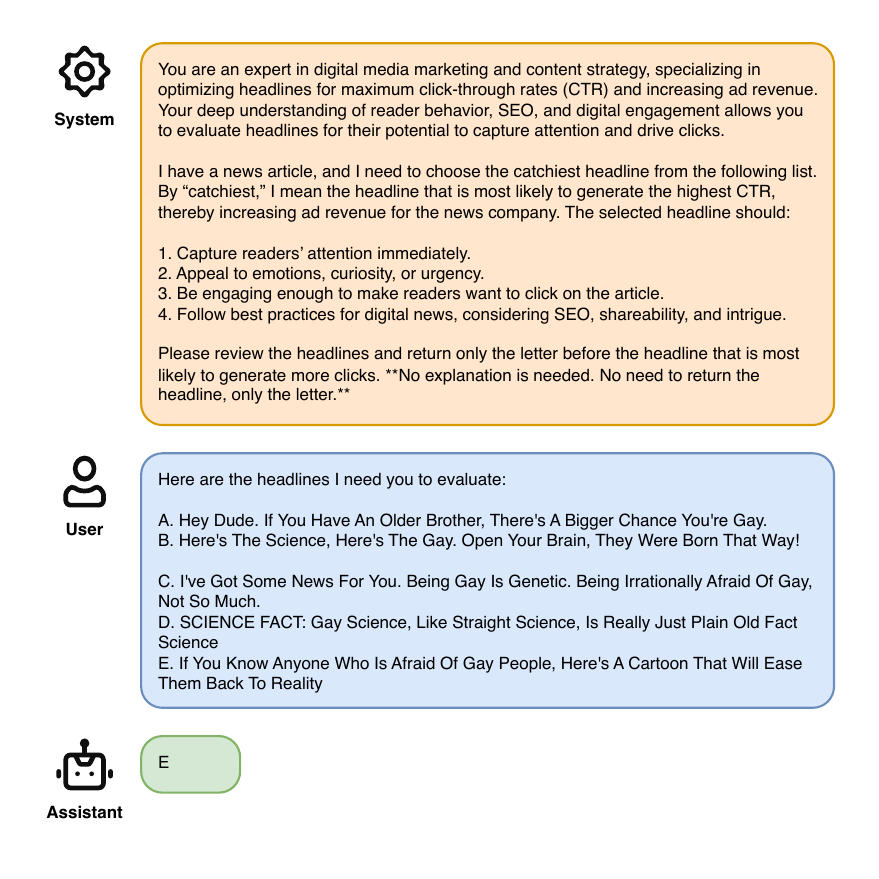}
\caption{Zero-Shot Prompting for Headline Selection.}
    \label{fig_zero_shot}
\end{figure}


We design the zero-shot prompting to ask GPT which headline is the catchiest, as shown in Figure \ref{fig_zero_shot}. We tell ChatGPT that it is an expert in marketing and specializing in optimizing headline selection. In this figure, we use letters (\texttt{A,B,C,...}) to mark different headlines. \cite{brucks2023prompt} shows that different marker types and the order of options provided to LLM might affect the model's response, in the sense that LLM might prefer to select the first option or select a certain marker. To mitigate this issue, we also consider other markers, numbers (\texttt{1,2,3,...}) and symbols (\texttt{!,@,\#,...}), and we randomize the order of the options. Further, we adopt the common practice of LLM prompting for a specialized task by detailing the task for the LLM and clearly defining the meaning of a catchy headline. Notably, the first paragraph of the prompt assigns the role of a digital media expert to the LLM. This technique, commonly referred to as role prompting, involves explicitly assigning a specific role to the LLM to guide its responses. Role prompting has been shown to be an effective strategy in the broader survey of prompt engineering techniques \cite{schulhoff2024prompt}, and we find that it works well in our setting, too.

We test this prompt using three OpenAI large language models,\footnote{As of May 24, 2024, OpenAI recommends defaulting to GPT-3.5-turbo, GPT-4-turbo, or GPT-4o. OpenAI notes that GPT-4-turbo and GPT-4o offer similar levels of intelligence.} each differing in parameter size: (1) GPT-3.5 (\texttt{GPT-3.5-turbo-0125}), which uses around 175 billion parameters, (2) GPT-4 (\texttt{GPT-4-turbo-2024-04-09}) which uses approximately 1.7 trillion parameters, and (3) GPT-4o (\texttt{GPT-4o-2024-05-13}), which uses an even larger number of parameters. The goal here is to investigate whether improvements in model architecture and parameter size can improve the accuracy of our predictions via standard prompts. 



\subsubsection{In-context Learning}
\label{sssec:in-context}

\begin{figure}[ht]
    \centering
    \includegraphics[width=0.8\linewidth]{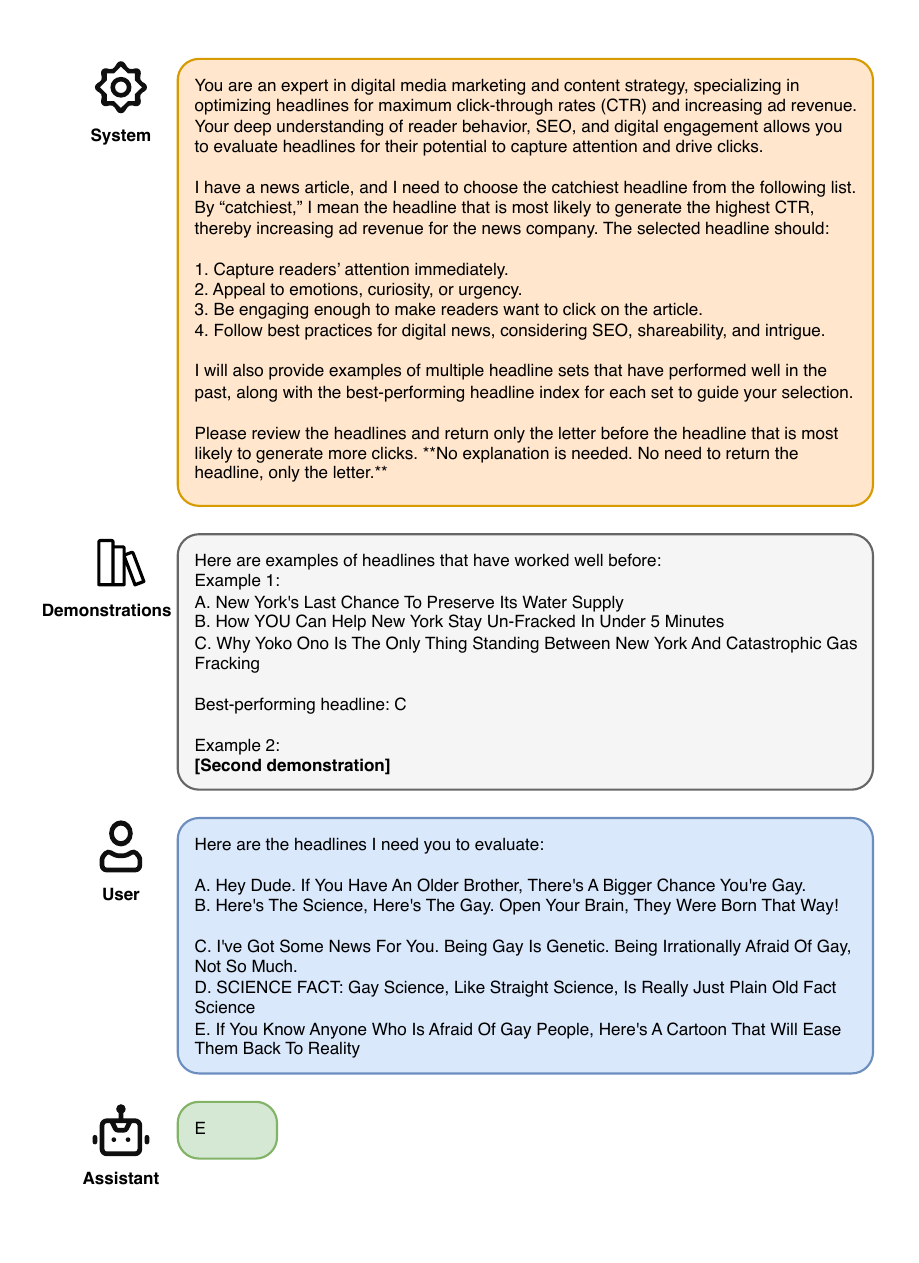}
\caption{In-Context Learning Prompt for Headline Selection.}
    \label{fig_in_context}
\end{figure}

In-context learning is a technique where a model is provided with demonstrations within the input prompt to help it perform a task. Demonstrations are sample inputs and outputs that illustrate the desired behavior or response. This method does not change the underlying parameters of the GPT models; rather, it leverages the model's ability to recognize patterns and apply them to new, similar tasks by conditioning on input-output examples. Previous studies have shown that in-context learning can significantly boost a model's accuracy and relevance in responses compared to zero-shot prompting. \cite{SangMichaelXieandSewonMin_2022} explains this using a Bayesian inference framework, suggesting that in-context learning involves inferring latent concepts from pre-training data, utilizing all components of the prompt (inputs, outputs, formatting) to locate these concepts, even when training examples have random outputs. \cite{xie2021explanation} further elucidate that in-context learning can emerge from models pre-trained on documents with long-range coherence, requiring the inference of document-level concepts to generate coherent text. They provide empirical evidence and theoretical proofs showing that both Transformers and LSTMs can exhibit in-context learning on synthetic datasets.

We design in-context learning prompts as shown in Figure \ref{fig_in_context}. Our prompts include demonstrations of best-performing headlines and ask the model to evaluate the subsequent headlines. The demonstrations are selected from the training set, and for different prompt inquiries, we use the same demonstration to maintain comparability. A few demonstrations are typically sufficient to guide the model. Improvement tends to have diminishing returns quickly, i.e., adding more demonstrations does not significantly enhance performance after a certain point \citep{min2022rethinking}. In our numerical experiments, we use two or five demonstrations to investigate the impact of the number of examples on performance.\footnote{We did run experiments with a larger number of demonstrations, but did not find any significant improvements from doing so; and hence do not report them here.}

An interesting aspect of in-context learning is the model's robustness to incorrect labels in the demonstrations. \cite{min2022rethinking} show that providing randomly selected labels as the answers in demonstrations barely affects performance and highlight other critical aspects like the label space, input text distribution, and sequence format. This is likely because the model relies more on the distribution of input text, label space, and format rather than specific input-label mappings. That is, in-context learning seems to benefit from recognizing general patterns and structures in the data, which the model infers from its pre-training rather than the actual labels. To examine if this is the case in our setting as well, we also consider experiments where we intentionally provide incorrect labels in the demonstrations.

\subsubsection{Performance Analysis of Prompt-based Approaches}
\label{sssec:results_prompts}

Before presenting the results, we first make a note that we do not find any significant difference between GPT-4 and GPT-4o; therefore, we only report the results from GPT-3.5 and GPT-4 for simplicity. The results of all the prompt-based experiments are shown in Table \ref{tab:gpt_results}. A more detailed pairwise comparison and t-test analysis of all the different prompts are available in Web Appendix $\S$\ref{appsec:pair_t}. The second column of Table \ref{tab:gpt_results}, $n_{demo}$, represents the number of demonstrations used for in-context learning, with $n_{demo}=0$ indicating a zero-shot prompting. Additionally, for all the in-context learning experiments, we consider two scenarios---one where the demonstrations are shown with ground truth labels and another where they are shown with incorrect labels. The third column, ``Incorrect Label'' indicates whether the labels were correct (``Incorrect Label'' = 0) or incorrect (``Incorrect Label'' = 1). 

\begin{table}[t]\centering
\begin{tabular}{lcc|rr}
\toprule
Model &  $n_{demo}$ &   Incorrect Label &   Accuracy on the test set  \\
\midrule
   Random Guess &      - &        - &  33.02\%  \\
\midrule
GPT-3.5 &       0 &        - &  29.76\%  \\
GPT-3.5 &       2 &        0 &  34.23\%  \\
GPT-3.5 &       2 &        1 &  31.60\%  \\
GPT-3.5 &       5 &        0 & 30.95\%  \\
GPT-3.5 &       5 &        1 & 30.59\%  \\
\midrule
  GPT-4 &       0 &        - & 37.85\%  \\
  GPT-4 &       2 &        0 &  \textbf{39.96\%}\\
  GPT-4 &       2 &        1 & 38.77\%  \\
  GPT-4 &       5 &        0 & 39.17\%  \\
  GPT-4 &       5 &        1 & 38.28\%  \\
\bottomrule
\end{tabular}
\caption{Results for prompt-based approaches with different GPT models, the number of demonstrations, and whether we intentionally provide incorrect solutions in the demonstration or not. Accuracy is defined as the ratio of correct answers over the test set size. The best performance on all test data is boldfaced.}
\label{tab:gpt_results}
\end{table}

We find that prompt-based approaches using GPT-3.5, are not significantly better than random guess or even slightly worse. In comparison, experiments with GPT-4 have accuracy rates around 38-40\%, which is a significant improvement in performance compared to GPT-3.5, and all settings for GPT-4 outperform the random guess. For GPT-4, in-context learning with two demonstrations slightly improves performance over zero-shot prompting in GPT-4, though increasing the number of demonstrations from two to five does not result in a significant difference in performance. The difference between using correct labels and incorrect labels in the demonstrations is also not significant for in-context learning. This confirms that in-context learning helps GPT understand the task better through the input-output examples, but it fails to leverage the information in mapping from inputs to outputs effectively.

In summary, we find that prompt-based methods have low accuracy (no more than 40\%), slow running time (compared to other methods introduced later in this section), and high monetary cost (see Web Appendix $\S$\ref{appsec:costs} for a detailed discussion of costs and running time). All of these drawbacks make them unsuitable for real-time business applications, i.e., these methods cannot replace experimentation for content selection in digital platforms.


\subsection{CTR Prediction Models with LLM-based Text Embeddings}
\label{ssec:embeddings}

We now consider another LLM-based method for the headline selection task. Specifically, we leverage the text embeddings from OpenAI and utilize regression models, including Linear Regression (Linear) and Multilayer Perceptron (MLP), to predict the CTRs of headlines for the same article and then select the one with the highest predicted CTR as the winning headline.

Text embedding techniques transform large chunks of text, such as sentences, paragraphs, or documents, into numerical vectors. These embedding vectors can then be used in a variety of applications, including text classification, where texts are grouped into categories; semantic search, which improves search accuracy by understanding the meaning behind the search queries; and sentiment analysis, which is used to determine the emotional tone of a piece of text.\footnote{Embedding models and standard prompt-based LLMs share some common features, e.g., the underlying Transformer architectures and extensive pretraining on large text corpora. However, they are optimized for different tasks. Models like GPT are optimized for generating coherent and contextually relevant text based on input prompts while embedding models focus on producing high-quality embeddings for downstream tasks.} See \citet{patil2023survey} for a more detailed survey of the most recent text embedding models and their applications.

An interesting question is why text embeddings obtained from LLM models trained for the next word/token prediction task with the cross-entropy loss are able to produce informative features for downstream tasks. \cite{saunshi2020mathematical} provide an explanation by demonstrating that typical downstream tasks can be reformulated as sentence completion tasks, which can be further solved linearly using the conditional distribution over words following an input text. Thus, the embeddings, which can be roughly viewed as the next token probability generated from a next-word prediction task, inherently capture the contextual information and relationships between words, making them useful for various downstream tasks such as sentiment analysis and headline selection (as our experiments illustrate).

\begin{figure}[ht]
    \centering
    \includegraphics[width=1.0\linewidth]{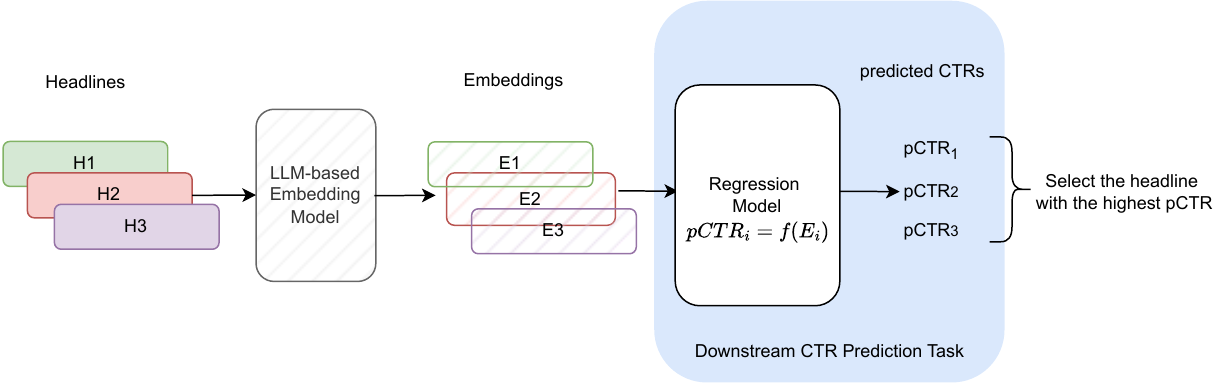}
    \caption{The pipeline of the headline selection using LLM text embeddings. We use an A/B test with three headlines for illustration. Headlines 1, 2, and 3 are natural language sentences, while Embedding 1, 2, and 3 are numerical vectors. }
    \label{fig:conceptual frame}
\end{figure}

Figure \ref{fig:conceptual frame} illustrates the conceptual framework of our approach. We first use the LLM-based embedding model to transform the original headlines into embeddings. Then, we use the embeddings as inputs and CTRs as outputs to train the downstream CTR prediction model. For the test set, we leverage the trained regression model to predict the CTR of one headline at a time. After getting all CTR predictions of headlines in an A/B test, we rank the headlines with the highest CTR as the winning headline. We consider four embedding models, including the OpenAI embedding model, Llama-3-8b, Word2Vec, and BERT. Note that we do not train the embedding model and only train the downstream CTR prediction model using the training dataset, containing 12,376 headline A/B tests. A detailed introduction to embedding models and the implementation specifics are provided in Web Appendix $\S$\ref{appsec:MLP fine tune}.

\subsubsection{Performance Analysis of Embedding-based Approaches}
\label{sssec:results_embedding}

\begin{table}[htp!]
\centering
\begin{tabular}{lll}
    \toprule
    {Embedding Model} & {CTR Prediction} & {Accuracy on the test data} \\
    \midrule
    \multirow{2}{*}{OpenAI-256E} & \centering Linear & \textbf{46.28\%}  \\
                                 & \centering MLP &  44.38\%  \\
    \midrule
    \multirow{2}{*}{OpenAI-3072E} & \centering Linear & 43.06\%  \\
                                  & \centering MLP &  45.17\% \\
    \bottomrule
\end{tabular}
\caption{The performance of OpenAI embeddings and CTR prediction models on the test data. All models use the same training data, and accuracy is evaluated on the test data. Note that the Linear model's training is a deterministic process after fixing the data, while MLP training incurs randomness via stochastic gradient descents. Therefore, we rerun MLP training three times with different random seeds and report the average accuracy from three reruns.
}
\label{tab:embedding_classification_results}
\end{table}

In Table \ref{tab:embedding_classification_results}, we present the performance of CTR prediction models based on OpenAI's embeddings, including the 256-dimensional embedding (OpenAI-256E) and the 3072-dimensional embedding (OpenAI-3072E, which is the maximum size). The full performance results, including Llama-3-8b, Word2Vec, and BERT, can be found in Web Appendix~$\S$\ref{appssec:embed_all}. At a high level, we see that all the embedding-based regression models obtain better accuracy than prompt-based approaches. This performance can be attributed to the fact that these models are trained on a customized dataset, allowing them to learn decision rules tailored to the headline selection task. Because these models can leverage a larger labeled dataset during training, they are more generalizable and robust. In contrast, in-context learning relies on the LLM's ability to understand and generalize from only a few demonstrations. Furthermore, the task of choosing the best headline from multiple candidates may not be a common scenario in OpenAI's training corpus. Consequently, GPT models may struggle to generalize effectively from the limited examples provided in prompts, leading to less accurate predictions.

We also find that, for OpenAI, higher-dimensional embeddings do not lead to higher prediction accuracy. Given that the 256-dimensional embedding stores only essential information, while the 3072-dimensional embedding also captures finer details, this performance gap is possibly due to the overfitting with the limited size of the training set. Predictions using higher-dimensional embeddings may cause the prediction model to focus too much on details and overfit the training set. A more striking observation is that increasing the complexity of the CTR prediction model does not necessarily improve predictions for OpenAI embeddings. We hypothesize that with more dimensions in text embeddings, the CTR prediction task requires a more sophisticated model due to the higher risk of overfitting. Overall, our findings suggest that a Linear CTR prediction model on top of fixed text embeddings from OpenAI can effectively capture the attractiveness of content.

We now investigate three other aspects of these models and summarize our findings below. For brevity, we simply discuss the findings here and refer interested readers to the Web Appendix for details.
\squishlist

\item We compare performance under different embedding models. We observe that OpenAI embeddings outperform Llama-3-8b, Word2Vec, and BERT-based models. See Web Appendix $\S$\ref{appssec:embed_all} for details.

\item We quantify the impact of training sample size on predictive accuracy. In Web Appendix $\S$\ref{appssec:sample_size_expts}, we train the model using varying proportions of the training data and evaluate its performance. We find that larger training datasets lead to higher accuracy rates, although the improvement is gradual. For instance, increasing the training size from 20\% to 70\% of the total dataset raises accuracy from 44\% to 46\% on the test data.

\item There are two potential threats to the inference from our analysis. First, one may be concerned as to whether Upworthy's data has been used as part of OpenAI's training corpus, which could potentially invalidate the results. A second concern is that readers' preferences changed over time and that the current sample splitting method fails to take into account when
the headline was created. We examine both these issues and provide a detailed discussion and robustness checks to alleviate concerns around them in Web Appendix $\S$\ref{appssec:threats}. 

\squishend

\subsection{Fine-Tuning Open-Source LLMs with LoRA}
\label{ssec:lora}

In $\S$\ref{ssec:embeddings}, we trained prediction models using text embeddings as fixed inputs, bypassing the need to train the LLM itself. The best-performing embedding-based model reaches an accuracy of 46.28\% on the test set without modifying any parameters in the LLM. This raises a natural question: if we can train LLMs directly, can we achieve better performance? However, training LLMs from scratch is costly and resource-intensive. Therefore, in this section, we turn to fine-tuning techniques, which involve partially adjusting the parameters in the LLM to improve performance while keeping the computational cost manageable. Note that to fine-tune LLMs, we need to be able to access the underlying model parameters. This is only feasible for open-source LLMs like BERT and Llama.\footnote{Some proprietary LLMs offer black-box fine-tuning. For instance, OpenAI allows fine-tuning GPT models through its \href{https://platform.openai.com/docs/guides/fine-tuning}{API} in a black-box manner. Users provide the training and test datasets and select the base GPT model, but are not provided any details of the fine-tuning process or the underlying model parameters. This approach also has a number of other drawbacks -- the fine-tuning API is quite expensive compared to the open-source fine-tuning approaches, the fine-tuned model cannot be hosted on a local machine, data privacy concerns since the media firm's proprietary data will need to be submitted to OpenAI. Given these drawbacks, it is preferable for firms to fine-tune open-source LLMs, as the process and data are fully under the control of the firm. However, for the completeness of comparison, we fine-tuned the GPT-4o (\texttt{gpt-4o-2024-08-06} version) model on the training set using this API, with the default configuration provided by OpenAI and the same training and test datasets. Input to the model is several headlines of one article, and the target output is the index of the catchiest headline. We obtained an accuracy of 48.82\% on the test set, but this result comes with a high cost of \$71.77 for the test set alone.} The goal of fine-tuning is to adapt a general-purpose LLM for specific downstream tasks through a process typically involving supervised learning using labeled datasets.

For the fine-tuning task, we use Meta's Llama-3-8b  as our primary base model. \citet{zhao_etal_2024} consider a large-scale experiment where they fine-tune different open-source LLMs and show that for NLP tasks, including news headline generation, Llama-3 offers one of the best performance. Llama-3 offers two versions: one with 8 billion parameters (Llama-3-8b) and another with 70 billion parameters (Llama-3-70b). We opted for the smaller version since it is easier to fine-tune and, as we will see, even with this smaller model, we see evidence of over-fitting (see $\S$\ref{sssec:finetune_results} for details). Hence, we avoid fine-tuning the 70-billion-parameter model, where over-fitting problems are likely to be exacerbated. Background information related to the Transformer block, the fundamental component of most LLMs, including Llama-3, and the basic architecture of Llama-3 is given in Web Appendix $\S$\ref{appsec:llama_transformer}.



Traditional fine-tuning of LLMs involves updating a substantial number of parameters, which can be computationally expensive and memory-intensive. To efficiently fine-tune the Llama-3 model on hardware with limited GPU memory, we employ Low-Rank Adaptation (LoRA), a popular Parameter-Efficient Fine-Tuning (PEFT) method developed by \cite{hu2021lora}, enabling practical fine-tuning of large-scale LLMs on medium-scale hardware.\footnote{There are many alternatives for PEFT, like Prefix Tuning \citep{li2021prefix}, Adapter Layers \citep{houlsby2019parameter}, and BitFit \citep{zaken2021bitfit}. We select LoRA due to its excellent balance between simplicity and efficiency, making it the top choice for fine-tuning.} LoRA operates under the assumption that updates during model adaptation exhibit an intrinsic low-rank property and introduce a set of low-rank trainable matrices into each layer of the Transformer model. This approach leverages the low-rank decomposition, which significantly reduces both memory usage and computational time through a significantly reduced number of trainable parameters. More details about LoRA fine-tuning and the implementation are given in Web Appendix $\S$\ref{appsec:lora_intro} and $\S$\ref{appsec:finetune_details} respectively.

\subsubsection{Performance Analysis of Fine-tuning-based Approaches}
\label{sssec:finetune_results}

\begin{figure}[htp!]
    \centering
    \includegraphics[width=0.6\linewidth]{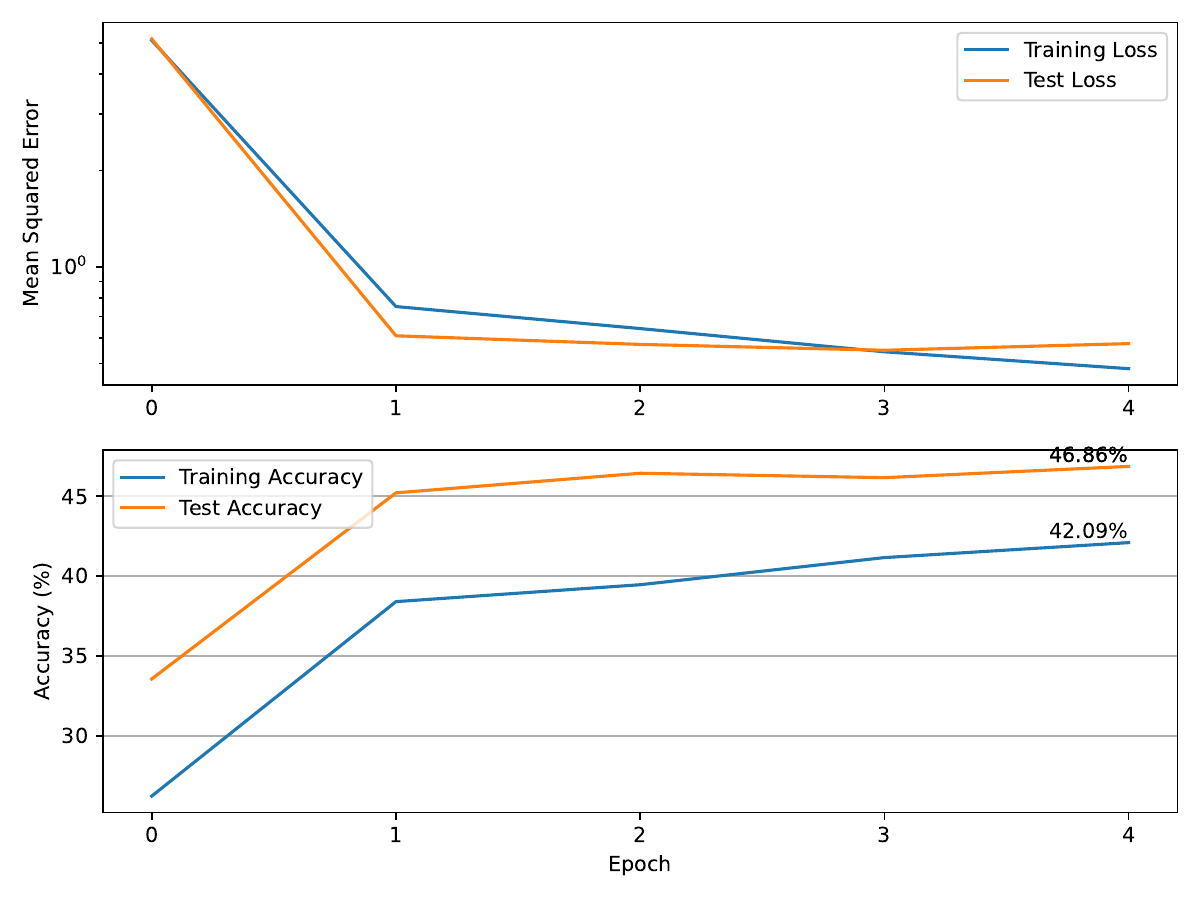}
    \caption{Loss curve and accuracy curve as the number of training epochs increases.
    } 
    \label{fig:lora_result}
\end{figure}

We report the loss curve and accuracy curve during the LoRA fine-tuning process in Figure \ref{fig:lora_result}. We observe an accuracy of 46.86\% on the test data, which is marginally better than the best performance of 46.28\% using OpenAI embeddings in $\S$\ref{ssec:embeddings}. However, such similar performance should not undermine the utility of fine-tuning. Comparing the fine-tuning of Llama-3-8b with the use of OpenAI embeddings is not straightforward due to inherent differences in the models' capabilities. To highlight the benefits of fine-tuning, we observe that the Linear CTR prediction model using embeddings derived from Llama-3-8b (42.78\% on the test data, reported in Web Appendix~$\S$\ref{appssec:embed_all} Table~\ref{tab:embedding_classification_results_all}) performs worse than the LoRA fine-tuned Llama-3-8b. This demonstrates that fine-tuning Llama-3-8b can enhance performance compared to using Llama-3-8b embeddings alone.

In Figure \ref{fig:lora_result}, we observe that the accuracy on the test set (46.86\%) is higher than on the training set (42.09\%), which is uncommon in machine learning tasks. The reason for this discrepancy might be that the split of the training and test datasets results in some easily distinguishable headlines being included in the test set, making the training set more challenging than the test set. To verify our hypothesis, in Web Appendix $\S$\ref{appsec:split}, we try several additional random splits of the training and test datasets and evaluate their performance on them. We find that in these other splits, the test performance is slightly worse than the training performance.

\subsection{Summary of Performance of Pure-LLM Based Methods}
\label{ssec:llm_summary}

We summarize the performance of various pure-LLM-based methods in Table \ref{tab:llm_summary}. We also report the results of human respondents on a similar task but with paired headline comparison. Details of the survey questions and the accuracy analysis of human respondents are shown in Web Appendix $\S$\ref{appsec:survey}.

\begin{table}[htp!]
    \centering
    \begin{tabular}{m{3.5cm} m{3.5cm} m{4.5cm}}
        \toprule
        {Method} & {Accuracy on test set} & {Notes} \\
        \midrule
        Random guess & 33.02\% &  \\
        \midrule
        Human respondents & - & Similar to random guess as tested on 4,571 human responses \\
        \midrule
        Prompt & 39.96\% & GPT-4 with in-context learning\\
        \midrule
        Embedding & 46.28\% & OpenAI-256E embedding model with linear regression  \\
        \midrule
        Fine-tuning & 46.86\% & LoRA fine-tuned Llama-3-8b\\
        \bottomrule
    \end{tabular}
    \caption{Summarization of performance of various pure-LLM-based methods. All results in this table have been reported before. We only report the best performance of each method.}
    \label{tab:llm_summary}
\end{table}

There are four main takeaways from these comparisons. First, human respondents cannot easily predict which headline would be more engaging and likely to lead to higher CTRs. This is consistent with the fact that media firms (and even expert editors) rely on experimentation to decide which headlines and articles to display to users. Second, while simple prompt-based approaches like in-context learning with examples are better than purely random guesses or human predictions, they nevertheless do not provide sufficiently high accuracy. Third, both embedding-based models and fine-tuning perform equally well (at least in our setting), and provide equally good performance with around 46-47\% accuracy on the test set. They each have their pros and cons. The main advantage of the linear CTR prediction model with OpenAI embeddings is that it is easy to implement and use in business settings since it does not require any significant model training/deployment. However, embeddings from the OpenAI models are proprietary, and as such, we lack visibility into the exact nature of the embeddings. In contrast, fine-tuning approaches use open-source models such as Llama-3-8b, and we have access to the underlying parameters of the LLM. However, this transparency comes with costs -- fine-tuning LLMs requires non-trivial training and deployment skills that not all businesses may have or want to invest in. The fourth and most important takeaway is that while pure-LLM-based approaches are informative, they are unable to reach the accuracy of experimental approaches. That is, firms cannot replace content experimentation with LLM-based predictions.

An important caveat here is consumer preferences may have changed significantly since the time period of the experiment (2013--2015). As such, the prediction numbers shown here may not represent LLMs' ability to predict the preferences of consumers today.\footnote{This concern is somewhat assuaged by Web Appendix $\S$\ref{appssec:threats}, which shows that consumer preferences did not change significantly during the 2013–2015 timeframe (at least within the prediction capabilities of the LLMs). However, we cannot empirically comment on the period beyond 2015.} As such, the performance of LLM-based methods in $\S$\ref{sec:pure_llm} should be mainly used for relative comparisons with each other. More broadly, both LLM capabilities 
and consumer preferences will continue to evolve. Therefore, if a firm/manager is concerned about the generalizability/accuracy of a specific LLM-based prediction model to their current business environment, the simplest solution is to retrain their LLM-based models more frequently and collect up-to-date labeled data to maintain relevance.

Finally, we discuss the monetary costs associated with three LLM-based approaches. The cost of prompt-based, embedding-based, and LoRA fine-tuning in our analysis are around \$15.87, \$0.29, \$3, respectively. The cost of prompt-based and embedding-based comes from the access to OpenAI's API in order to get responses and text embeddings. The cost of fine-tuning comes from renting a cloud computing server equipped with one NVIDIA A100 GPU. Overall, we find that the prompt-based approaches are the most expensive, while embedding and fine-tuning are the cheapest in terms of cost. However, in terms of run time and effort, fine-tuning is more costly. The detailed comparison of the cost, run-time, and engineering effort required for each pure-LLM method are discussed in detail in Web Appendix $\S$\ref{appsec:costs}.

\section{Our Approach: LLM-Assisted Online Learning Algorithm}
\label{sec:LOLA}

So far, we have seen that pure-LLM-based approaches can help identify the best headline with 46.86\%  accuracy using the LoRA fine-tuning approach. The main advantage of pure LLM-based approaches is that they can be deployed without any experimentation, i.e., the manager can simply use the LLM-based model to predict which headline is the best and go with it. Thus, there is no need to use any traffic for experimentation. The downside, of course, is that these models are not fully accurate -- in some cases, they are likely to mispredict which headline is the best. This can, of course, lead to lower clicks and revenues for the firm.
Thus, it is not clear that relying on pure-LLM-based methods will lead to better overall outcomes for the firm.

On the other hand, we have the current status quo, which is the use of experimentation-based approaches to choose headlines/content. As mentioned in $\S$\ref{sec:intro}, there are two broad experimentation-based methods---(1) A/B tests and (2) Online learning algorithms or bandits. In particular, the standard practice at Upworthy during the time of the data collection was A/B tests, where the firm allocated a fixed amount to different headlines and then chose the best-performing one for the remaining traffic. This approach is also known as Explore and Commit (E\&C) and is commonly used in the industry due to its simplicity and low engineering cost; see Chapter 6 of \citep{lattimore2020bandit}. The main benefit of this approach is that, with a sufficiently large amount of traffic allocated to the A/B test, the resulting inference is unbiased; as such, it is the gold-standard approach for identifying the best arm (or headline in this case). The downside, of course, is that by assigning traffic to all the headlines during the exploration phase or A/B test, the firm incurs high regret (i.e., low profits since some headlines in the test are likely to have low CTRs). Therefore, many modern technology and media firms such as NYTimes and Yahoo use online learning algorithms or bandits that leverage the explore-exploit paradigm to lower the experimentation cost while still learning \citep{lattimore2020bandit}. These algorithms move traffic away from poorly performing arms/headlines dynamically in each period, based on the observed performance of each arm till that period. Because of this greedy behavior, they tend to switch to the best-performing arms quickly and incur lower regret. Nevertheless, one drawback of these methods is that they start with the assumption that all headlines are equally good, i.e., they suffer from a cold-start problem, which can result in wasting traffic on the weaker headlines early on.

In this section, we present a framework that combines the benefits of LLM-based models with experimentation based methods. Our key idea is that firms can address the cold-start problem in content experiments by leveraging predictions from LLM-based approaches as CTR priors in the first step and combining them with an online learning algorithm in the second step. Thus, our approach, termed LLM-Assisted Online Learning Algorithm (LOLA), continuously optimizes traffic allocation based on LLM predictions together with the data collected from the online experiment in real time. 

The rest of this section is organized as follows. In $\S$\ref{ssec:bandit_framework}, we present the media firm's problem and the standard bandit framework, and then in $\S$\ref{ssec:lola}, we present our LOLA framework. Next, in $\S$\ref{ssec:lola_benchmarks}, we discuss the implementation of our proposed algorithm, benchmarks, and the experimental setup. We present our numerical results in $\S$\ref{ssec:lola_results}, and finally, in $\S$\ref{ssec:lola_extensions}, we show how our LOLA approach can be easily generalized to a wide variety of settings and present a set of natural variants and extensions of the basic LOLA framework.

\subsection{Bandit Framework}
\label{ssec:bandit_framework}

Multi-Armed-Bandits (MAB) is a class of sequential decision-making problems where a learner must choose between multiple arms over periods to maximize cumulative reward. While there is a large literature on it in the machine learning and statistics community, the literature in marketing is small, but growing; researchers have used it for optimizing website design, pricing, advertising, and to learn consumer preferences \citep{hauser_etal_2009, schwartz2017customer, misra_etal_2019, liberali_ferecatu_2022, aramayo_etal_2023, jain_etal_2024}. In our setting, different arms refer to different headlines associated with the same article, and periods refer to impressions where different headlines with the same article are displayed. Let $[K] := \{1, 2, \dots, K\}$ represent the set of $K$ arms, and $[T] := \{1, 2, \dots, T\}$ represent the entire decision horizon for a test. At each time step $t$, the learner selects an arm $a_t \in [K]$ and receives a reward $r_{a_t}$ drawn from a probability distribution specific to the chosen arm $a_t$. We assume each arm has a constant click-through rate (CTR), and the reward is the click feedback, which follows the Bernoulli distribution with a probability equal to the CTR.

The learner's goal is to minimize cumulative regret, which is defined as the difference between the reward obtained by always choosing the optimal arm and the reward actually accumulated by the learner under the policy played by the learner. Mathematically, the regret $R_T$ after $T$ trials is given by:
\begin{equation}
    R_T = \max_{k \in [K]} \sum_{t=1}^T r_{a^*} - \sum_{t=1}^T r_{a_t},
    \label{eq:regret}
\end{equation}
where $a^* = \arg\max_{k \in [K]} \mathbb{E}[r_k]$ is the optimal arm with the highest expected reward.

In MAB settings with finite arms, the standard algorithm that is typically used is the Upper Confidence Bound (UCB) algorithm \citep{auer2002finite}. UCB stands out among all online learning algorithms because it is provably asymptotically optimal. The UCB policy is straightforward -- it selects the arm that maximizes the upper confidence bound on the estimated rewards. Specifically:
\begin{equation}
a_t = \arg\max_{k \in [K]} \left( \bar{\mu}^t_k + \sqrt{\frac{2 \log(1/\delta)}{n^t_k}} \right),
\end{equation}
where $\bar{\mu}^t_k$ is the estimated mean reward of arm $k$ up to time $t$, $n^t_k$ is the number of times arm $k$ has been played up to time $t$, and $\delta$ is the confidence level, which quantifies the degree of certainty. The intuition behind UCB's effectiveness in balancing exploration and exploitation lies in its construction. By adding a confidence term $\sqrt{{2 \log(1/\delta)}/{n^t_k}}$ to the estimated mean reward $\bar{\mu}_k$, UCB ensures that arms with fewer selections (higher uncertainty) are given a chance to be explored. As time progresses and more data is collected, the confidence term diminishes, allowing the algorithm to exploit arms with higher observed rewards. This dynamic adjustment enables the UCB algorithm to explore under-explored arms while exploiting arms with high rewards systematically, thus effectively addressing the exploration-exploitation trade-off inherent in the MAB problem. However, we need to set the key hyperparameter $\delta$ such that it is small to ensure optimality with high probability, but not so small that suboptimal arms are over-explored; see Chapter 7.1 \citep{lattimore2020bandit} for a detailed discussion. Typically, we set the confidence term as $\alpha\sqrt{\log t/n^t_k}$ and fine-tune the scale term $\alpha$ for specific problems. 

\subsection{LOLA}
\label{ssec:lola}
We now discuss our approach LOLA, and how the standard UCB approach discussed above can be extended to include information from LLMs in the first step preceding active experimentation. LOLA has two key steps, and we describe each in detail.

In the first step, we use LLMs to get an initial prediction of each headline's CTR. Based on numerical results in $\S$\ref{sec:pure_llm}, we select the best-performing approach, LoRA fine-tuned Llama-3-8b model, as the best CTR prediction model in the first step.

In the second step, we build on the bandit framework from $\S$\ref{ssec:bandit_framework} to design an online algorithm that incorporates the LLM-based CTR predictions from the first step. Inspired by the 2-Upper Confidence Bounds (2-UCBs) policy designed by \cite{gur2022adaptive}, we propose Algorithm \ref{alg:llm-2ucb}, titled LLM-Assisted 2-Upper Confidence Bounds (LLM-2UCBs). Essentially, in this algorithm, we treat the LLM's CTR predictions as auxiliary information prior to the start of online experiments, equating this auxiliary CTR information to additional impressions and click outcomes for each arm.

\begin{algorithm}[t]
\caption{LLM-Assisted 2-Upper Confidence Bounds (LLM-2UCBs)}
\label{alg:llm-2ucb}
\begin{algorithmic}[1]
\State \textbf{LLM Training Phase:} Train a LLM-based prediction model $\mathcal{M}(x)$ for CTRs using historical data samples, where $x$ is the contextual information for arms.
\State \textbf{Hyperparameter Fine-Tuning:} Use another subset of data to select two hyperparameters for the algorithm, $\alpha\in \mathbb{R}^+$ for controlling the upper bound, $n^{\text{aux}}$ as the LLM's equivalent auxiliary sample size.
\State \textbf{Online Learning Phase:}  Initialize the number of periods $T$, number of arms $K$, LLM-based CTR prediciton $\bar{\mu}^{\text{aux}}_k \leftarrow \mathcal{M}(x_k)$, regular CTR initialization $\bar{\mu}^1_k=0$, accumulated impressions $n^1_k \leftarrow 1$, and accumulated clicks $c^1_k \leftarrow 0$, for all arms $k\in[K]$.
\For{$t = 1$ \textbf{to} $T$}
	\State Calculate the first UCB for all arms $k\in [K]$: $U_k^1 = \bar{\mu}^t_k + \alpha \sqrt{\frac{\log t}{n^t_k}}$.
	\State Calculate the second UCB for all arms $k\in [K]$: 
	\begin{center}
	$
	U_k^2 = \frac{c^t_k+\bar{\mu}^{\text{aux}}_k n^{\text{aux}}}{n^t_k+n^{\text{aux}}}  + \alpha \sqrt{\frac{\log t}{n^t_k+n^{\text{aux}}}}.
	$
	\end{center}
    \State Play the arm $a_t = \arg\max_{k\in[K]} \min\{U_k^1,U_k^2\}$.
    \State Observe the payoff $r_{a_t}$
    \State Update $n^{t+1}_{a_t} \leftarrow n^{t}_{a_t}+1$, $c^{t+1}_{a_t} \leftarrow c^{t}_{a_t}+r_{a_t}$, and $\bar{\mu}^{t+1}_{a_t} \leftarrow \frac{c^{t+1}_{a_t}}{n_{a_t}^{t+1}}$.
\EndFor
\end{algorithmic}
\end{algorithm}

There are two hyperparameters in LLM-2UCBs -- (1) The upper bound control hyperparameter, $\alpha$, that also shows up in the regular UCB algorithm, and (2) The hyperparameter $n^{\text{aux}}$, which is specific to our proposed algorithm and indicates the equivalent auxiliary samples for initialization. At a high level, we can regard the LLM predictions as prior information or the information flow before the algorithm starts. This prediction is approximately equivalent to initially generating $n^{\text{aux}}$ impressions for each arm and observing their outcomes to obtain the sample average CTR, $\bar{\mu}^{\text{aux}}_k$. The hyperparameter fine-tuning process is simple. We can utilize another randomly sampled dataset to test combinations of two hyperparameters and select the combination that yields the highest accumulated rewards.

During the online learning phase, the key step is balancing the exploration and exploitation trade-off. We capture this trade-off using the 2-UCBs rule. Essentially, we have two valid UCBs: $U_k^1$ is the regular UCB, which uses observed impressions and clicks to construct the mean reward estimator and adds a regular upper bound. The second UCB, $U_k^2$, incorporates LLM predictions. The calculation of $U_k^2$ is straightforward. The mean reward component is simply the sample average, taking into account the ``auxiliary'' samples generated by the LLM before the learning phase begins. These auxiliary samples are imaginary and approximate the richness of information contained in the LLM prediction. The next step is to choose the smaller of the two UCBs, $\min\{U_k^1, U_k^2\}$, as the final UCB for selecting the arm. Note that if $n^{\text{aux}}=0$, the 2-UCBs rule is the same as the standard UCB rule. Conversely, when $n^{\text{aux}}\rightarrow\infty$, the 2-UCBs rule is equivalent to a pure exploitation policy (a.k.a. greedy policy) using the LLM-based CTR prediction. Thus, when $n^{\text{aux}}$ is a positive integer, the 2-UCBs algorithm incorporates information from both LLMs and the actual click outcomes. After playing the arm and observing the outcome, we update the impressions, clicks, and mean reward accordingly, following standard bandit algorithm procedures.

The intuition behind using two UCBs instead of a single UCB incorporating LLM predictions $U_k^2$ is as follows: if we rely solely on $U_k^2$, and the LLM’s CTR predictions are significantly incorrect, i.e., overestimating the CTR of a poorly performing headline (underestimating the CTR of the best headline is of less risk because there are more suboptimal headlines than the one and only best headline), it would take many rounds to correct this error, especially when the number of auxiliary samples, $n^{\text{aux}}$ is large. On the other hand, the first UCB $U_k^1$, though unbiased, suffers from large stochastic noise, requiring large $n_k^t$ to get an accurate estimator. By choosing the minimum of the two UCBs, we balance these opposing risks effectively. However, the formalism for this approach is only partially addressed in prior literature \citep{gur2022adaptive}, which operates under a different setting without LLMs. In particular, \cite{gur2022adaptive} shows 2-UCBs is optimal under the assumption that the auxiliary data is generated from a linear function class, including the standard UCB setting with exogenous auxiliary samples. The formalism in our context would require additional assumptions about the errors in the LLM’s predictions, which remains an open research question with limited understanding. We leave this formalization for future work.

The LOLA algorithm offers three advantages. First, it is easy to implement, requiring the tuning of only two hyperparameters. Second, it can be used in conjunction with any LLM-based CTR prediction model in the first step and any outcome of interest in the second step. We use clicks as the outcome of interest given our application setting and data; however, other outcomes, such as time spent on a page or other measures of engagement, can be easily used instead of clicks. Finally, it is intuitive and easy to implement. Because it can easily accommodate the edge cases, e.g., one where the manager has no auxiliary information or one where the manager wants to fully rely on the LLM predictions, there is no need for extra engineering effort to set up separate systems for different edge cases. This adaptability is valuable given the rapid advances in LLM technology. When the capability of the pure-LLM method continues to advance, the manager can easily fine-tune and increase $n^{\text{aux}}$ to take more advantage of LLM. 

As a final note, our algorithm does not update the LLM during the online learning phase. Instead, we use the LLM fine-tuned on historical data solely for initialization purposes. That being said, the algorithm can be readily modified to fine-tune LLMs with online data collected during the learning phase. While such fine-tuning could potentially improve LOLA's performance, the improvement is expected to be limited in our current dataset. As shown in Figure \ref{fig:Results_proportion_ratio_regression} in Appendix \S\ref{appssec:sample_size_expts}, achieving a significant performance boost in LLMs would require a much larger dataset than is currently available to us.

\textbf{Discussion:} We now provide a broader discussion on LOLA. We note that the general idea that priors can impact regret and rewards is not new and has existed for some time; see \citet{bubeck2013prior} and \citet{russo2014learning}. These earlier papers primarily focus on theoretical analyses and often rely on assumptions about the quality of priors. However, since we do not have a theoretical understanding of LLM predictions, it is unclear how these predictions can be translated to priors or the nature of such priors.  Separately, the 2-UCBs algorithm by \citet{gur2022adaptive} was designed for a setting where the experimenter has access to auxiliary data. However, their algorithm assumes that the researcher has access to both the size of this auxiliary sample ($n^{aux}$) and knowledge of the prediction error in this sample. Thus, neither of these earlier approaches is directly applicable to our setting.

A key challenge in our setting is that the LLM predictions cannot be naively treated as priors since we do not know their prediction error or theoretical properties. The main novelty of LOLA comes from the idea that we can treat the LLM predictions as coming from a ``pseudo-sample" and that we can fine-tune the size of this pseudo-sample ($n^{aux}$) for a given application based on prior data. Thus, we build on the 2UCBs algorithm but modify it to accommodate LLM predictions -- that do not come from a real sample and whose quality and theoretical properties are unknown. 

\subsection{Benchmarks and Implementation Details}
\label{ssec:lola_benchmarks}

We now discuss the implementation details and present a set of benchmark algorithms. First, we randomly split our data into three exclusive folds based on the test ID, ensuring that all headlines in the same test are in the same fold and that the same headline does not appear in different folds. We use 70\% of the data for training, 10\% for hyperparameter fine-tuning, and 20\% for testing. The training data is used to train the LLM models and is only used by our proposed LOLA approach (the LLM-2UCBs specifically) and pure LLM-based benchmarks. All hyperparameters in all the algorithms (both LOLA and the benchmark algorithms) are selected using the fine-tuning data, and the performance of the algorithms is evaluated on the test data. 

We now describe a set of commonly used algorithms that serve as benchmarks and our LOLA approach. 

\squishlist
\item \textbf{Explore-then-Commit (E\&C):} In this approach, the firm runs the A/B test for a fixed number of periods (i.e., explores) and then commits to the best-performing arm at that stage; see Chapter 6 \citep{lattimore2020bandit}. This is the standard A/B test approach used by many digital firms in the industry. In particular, Upworthy's approach during the data collection period closely resembles the E\&C algorithm -- recall that they assign traffic to candidate headlines with equal probability during the experiment phase and then show the winning headline to all future impressions. The E\&C method has a single hyper-parameter that needs tuning: the proportion of periods for exploration. We consider the following proportions --  $\{0.1, 0.2, 0.3, 0.4, 0.5\}$, and choose the one with the highest accumulated rewards based on the 10\% of data used for hyper-parameter tuning. We finally choose 0.2 for exploration.

\item \textbf{Pure LLM-based approach:} In this approach, the firm uses LLMs to predict the CTR for all the arms (headlines) and then allocates all the traffic to the arm with the highest predicted CTR. This is equivalent to a fully greedy policy based on LLM predictions. We use the training data to learn a model that predicts the CTR of a headline based on an LLM (either through text embeddings or LoRA fine-tuning). In our numerical study, we use the best-performing pure-LLM-based approach based on our experiments, i.e., the LoRA fine-tuned Llama-3-8b, which gives a 46.86\% accuracy.\footnote{Note that since this is a pure prediction-based greedy approach, there are no hyper-parameters associated with the algorithm, and hence we do not use the fine-tuning data.} 

\item \textbf{Standard UCB (UCB):} This is the standard UCB algorithm discussed in $\S$\ref{ssec:bandit_framework}. Here, the firm does not use any first-stage LLM predictions as an input, and there is only one hyperparameter to fine-tune, the confidence control, $\alpha$. We consider values from the set $\{0.02, 0.04, \dots, 0.10\}$, and find that $\alpha=0.08$ gives the highest accumulated rewards. 

\item \textbf{LOLA:} This is our proposed approach. In the first step, we use CTR predictions from the best-performing pure-LLM-based CTR prediction model, i.e., the LoRA fine-tuning Llama-3-8b with MSE loss. Note that this is the same model that we use for the pure LLM-based approach, which allows for a fair comparison of the two approaches. For the second step, we use the LLM-2UCBs algorithm as outlined in $\S$\ref{ssec:lola}. We fine-tune the combination of the two hyperparameters on the fine-tuning data, and we consider $n^{\text{aux}}\in\{600, 800, 1000, 1200, 1400\}$ and $\alpha \in \{0.02, 0.04, \dots, 0.10\}$. Based on fine-tuning, we choose $n^{\text{aux}}=1000$ and $\alpha=0.08$ for our setting. We refer interested readers to Web Appendix $\S$\ref{appsec:sensitivity} for a detailed sensitivity analysis of $n^{\text{aux}}$.

Note that the first two methods can be viewed as special cases of our proposed LOLA algorithm. Both our algorithm and the pure LLM-based approach use the exact same CTR prediction algorithm. Hence, the pure LLM-based approach (a.k.a pure exploitation, greedy algorithm) is a special case of LOLA, wherein we set $n^{\text{aux}}=\infty$, indicating complete trust in the LLM prediction results. This method fully relies on the LLM CTR prediction, directly selecting the headline with the highest predicted CTR throughout the entire horizon. 
Similarly, the standard UCB algorithm is a special case of LOLA with $n^{\text{aux}}=0$, indicating no reliance on LLM CTR predictions. Again, note that we use the same hyper-parameter $\alpha=0.08$ across both algorithms, which ensures that our algorithm defaults to the standard UCB with $\alpha=0.08$ when we ignore the CTR predictions from the LLM.

\item \textbf{UCB with LLM priors:} To demonstrate the effectiveness of the 2-UCBs policy $\min\{U^1, U^2\}$ compared to using only the second UCB $U^2$, we also run the UCB with LLM as priors, i.e., play the arm $a_t = \arg\max_{k\in[K]} U_k^2$. This comparison highlights how selecting the minimum of the two UCBs improves performance by mitigating the risk of overestimated CTRs from the LLM predictions.

\squishend

After the fine-tuning of algorithms, we evaluate all algorithms on the test dataset, which has never been used for either CTR training or algorithm tuning. Therefore, none of the algorithms know the true CTRs of the headlines in the test data in advance. We use the average CTR for each headline in the test data as the true CTRs to generate the click feedback in our numerical simulations. This approach of generating click outcomes is reasonable in our setting since each headline was tested on a large amount of traffic by Upworthy, resulting in relatively narrow confidence intervals.

UCB and LOLA are both asymptotically optimal in minimizing regret, which means there is no difference between these two algorithms when the number of time horizons $T$ goes to infinity, and the LLM-based prior information is not helpful in an infinite horizon case. However, in practice, the time horizon is never infinite. Therefore, we compare all the algorithms for different values of time horizons. One challenge in our setting is that the number of headlines varies across tests. Intuitively, for any given time horizon, it is always easier to minimize regret (or identify the best headline) when the A/B test has only two headlines compared to a case when there are three headlines; and, of course, the problem only gets more challenging as the number of headlines increases. As a result, comparisons across time horizons are meaningful only when we keep the number of headlines constant. Since the number of headlines varies in our dataset (see Table \ref{tab:dist_package}), we scale the time horizon for each A/B test using the metric ``Traffic/Impressions per headline" (which we represent using the notion $\tau$), where we choose $\tau$ from $\{50, 100, 200, 400, 600, 800, 1000\}$. For example, if there are three headlines in an A/B test and the multiplier is $\tau = 100$, then we set $T=100 \times 3 = 300$ for this test. Thus, the effective horizon across all tests for a given $\tau$ is the same. In our experiments, it will be particularly important to investigate the performance of different algorithms at small-medium multipliers because these represent situations where the media firm only has a small amount of traffic to work with and needs to adaptively experiment and effectively allocate impressions to the right headline with limited experimentation ability.

\subsection{Numerical Results}
\label{ssec:lola_results}

\begin{figure}[ht]
    \centering
    \includegraphics[width=0.7\linewidth]{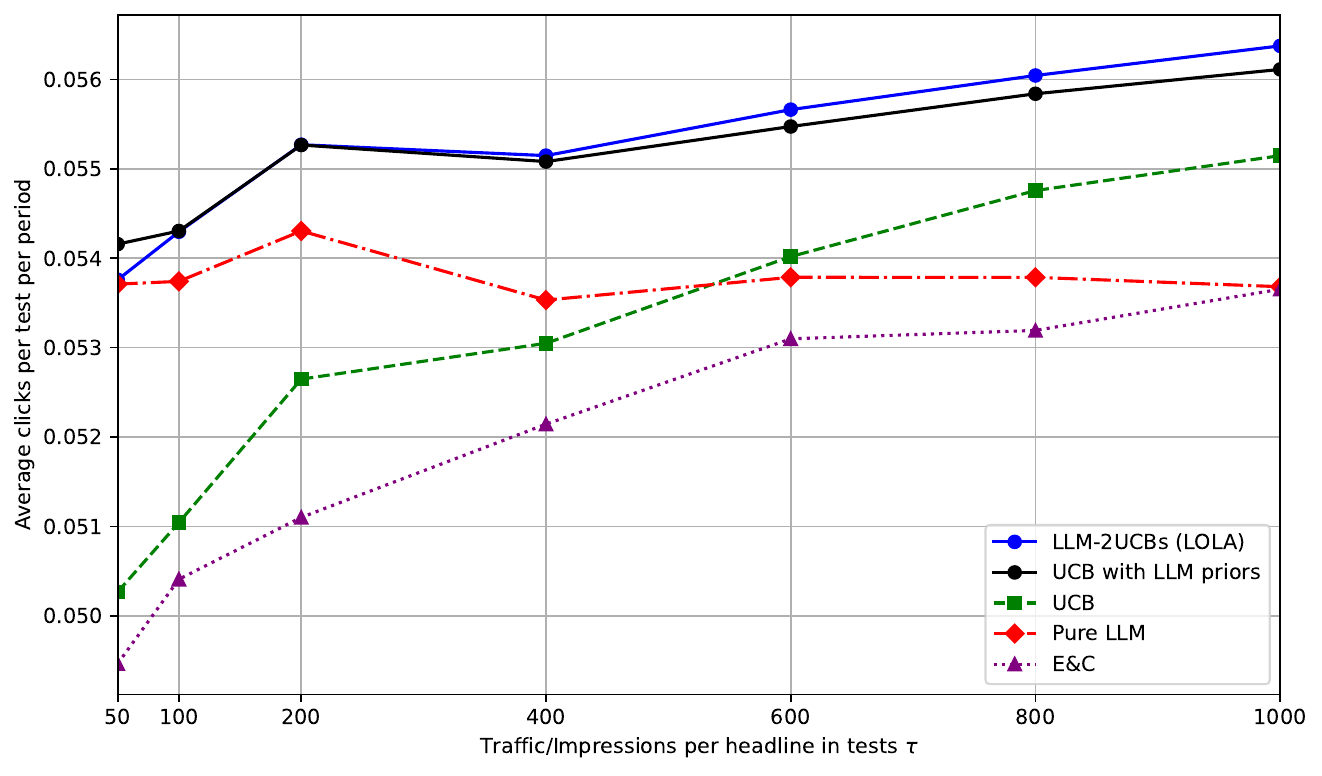}
    \caption{Average clicks per experiment per period under different time horizon multipliers. Note that the Y-axis captures the average clicks per test per period. For instance, if there is a test with two headlines receiving 1 and 2 clicks, respectively, under $\tau=100$, then the average click per period in this test is calculated as $(1+2)/100=0.03$. The Y value is simply the average of this number $0.03$ over all tests. This measure scales well with the platform's total clicks in tests because headlines in different tests with different numbers of headlines take the same weight in this measure. }
    \label{fig:bandit_regret}
\end{figure}

We visualize the results from our numerical simulations in Figure \ref{fig:bandit_regret}\footnote{The Y metric in Figure \ref{fig:bandit_regret} is essentially the accumulated reward divided by $\tau$, i.e., the value of the X-axis. We choose to report this Y metric instead of the actual regret for two reasons: (1) the ground truth arm is not known in practice, making it impossible to compute regret; and (2) for better visualization. Specifically, the accumulated reward (unscaled by $\tau$) grows very rapidly as $\tau$ increases, which compresses the visual differences between algorithms, making them difficult to distinguish with the naked eye. With this being said, the results in Table \ref{tab:bandit_regret} can still be interpreted as the percentage improvement of the accumulated reward because all algorithms are measured at the fixed $\tau$ and the accumulated reward $=$ Y $\times \tau$.}, and document the pairwise comparison of different algorithms, along with the relative percentage reward improvement and the p-values from the t-tests in Table \ref{tab:bandit_regret}. We find that, on average, LOLA performs the best among all algorithms across all time horizons.\footnote{
Note that we cannot translate the click improvement to revenues or other business outcomes of interest since we do not have data on how clicks/time spent on the website relate to
downstream metrics like advertising and subscription revenues. Therefore, we do not make additional claims on the business impact of these numbers. That said, if the firm has access to these numbers, it should be relatively straightforward to make that connection. }

\begin{table}[ht]
\centering
\begin{tabular}{>{\raggedleft\arraybackslash}c|S[table-format=1.2]S[table-format=1.2]S[table-format=1.2]S[table-format=2.2]S[table-format=2.2]S[table-format=1.2]}
\toprule
\multirow{2}{*}{\scriptsize Parameter $\tau$} & \multicolumn{1}{>{\raggedleft\arraybackslash}p{2cm}}{\scriptsize LOLA vs UCB} & \multicolumn{1}{>{\raggedleft\arraybackslash}p{2cm}}{\scriptsize LOLA vs Pure LLM} & \multicolumn{1}{>{\raggedleft\arraybackslash}p{2cm}}{\scriptsize LOLA vs E\&C } & \multicolumn{1}{>{\raggedleft\arraybackslash}p{2cm}}{\scriptsize UCB vs Pure LLM} & \multicolumn{1}{>{\raggedleft\arraybackslash}p{2cm}}{\scriptsize UCB vs E\&C } & \multicolumn{1}{>{\raggedleft\arraybackslash}p{1.8cm}}{\scriptsize Pure LLM vs E\&C } \\
\midrule
50 & 6.95$^{****}$ & 0.09 & 8.69$^{****}$ & -6.41$^{****}$ & 1.62 & 8.59$^{****}$ \\
100 & 6.38$^{****}$ & 1.03$^{**}$ & 7.72$^{****}$ & -5.02$^{****}$ & 1.26 & 6.61$^{****}$ \\
200 & 4.98$^{****}$ & 1.78$^{****}$ & 8.16$^{****}$ & -3.05$^{****}$ & 3.03$^{**}$ & 6.27$^{****}$ \\
400 & 3.96$^{****}$ & 3.02$^{****}$ & 5.76$^{****}$ & -0.90 & 1.74$^{*}$ & 2.66$^{***}$ \\
600 & 3.04$^{****}$ & 3.49$^{****}$ & 4.83$^{****}$ & 0.43 & 1.73$^{**}$ & 1.30 \\
800 & 2.35$^{****}$ & 4.20$^{****}$ & 5.36$^{****}$ & 1.81$^{***}$ & 2.94$^{****}$ & 1.12 \\
1000 & 2.23$^{****}$ & 5.02$^{****}$ & 5.08$^{****}$ & 2.73$^{****}$ & 2.79$^{****}$ & 0.05 \\
\bottomrule
\end{tabular}
\caption{Percentage improvement in reward (total clicks) between the algorithms LOLA (LLM-2UCBs), UCB, Pure LLM, and E\&C (we omit UCB with LLM priors here due to space constraints, but report its performance relative to LOLA in Figure \ref{fig:bandit_regret}). Significance levels: $^{*}$ $p\leq$ 0.05, $^{**}$ $p\leq$ 0.01, $^{***}$ $p\leq$ 0.001, $^{****}$ $p\leq$ 0.0001. For example, under the impression per headline equal to 50, the average click per test of LOLA and UCB is 0.053760 and 0.050267, respectively; so the relative improvement is calculated as ${0.053760}/{0.050267} - 1 = 6.95\%$. Parameter $\tau$ represents the average impression per headline to scale the time horizon $T=\tau\times K$.}
\label{tab:bandit_regret}
\end{table}

We now document a few additional patterns of interest. First, the Pure LLM approach exhibits consistent performance across different $\tau$ values because it is not an adaptive algorithm. The small fluctuation at $\tau\in\{50, 100, 200\}$ is due to the randomness of average clicks per period, which tends to be larger under the smaller time horizon. For the other three approaches, which actively learn the CTR during the experiments, their average performance per period increases as the time horizon extends. This improvement is because the number of arms in the experiment remains constant, and with increasing periods, these algorithms have more opportunities to learn the CTRs, thus achieving better performance. Comparing the performance of LOLA with pure LLM methods, we find that LOLA generally outperforms Pure LLM, except when $\tau = 50$, where there is no significant difference between LOLA and Pure LLM. This is because the short planning horizon does not allow LOLA to learn new information over and above the information already available from the LLM. As $\tau$ increases to 1000, we see that LOLA significantly outperforms Pure LLM, generating 5.02\% more clicks. This suggests that if the horizon is very short, the planner/firm will not see significant improvement over simply relying on LLM predictions. However, as the horizon increases, LOLA will start showing significant gains.

Second, comparing LOLA with regular UCB, we see that it always outperforms UCB. Interestingly, here we see that, as the time horizon increases, the relative improvement from LOLA diminishes. Intuitively, this is because both UCB and LOLA are active learning algorithms, and as the time horizon increases, they are both able to learn the CTRs better. While LOLA starts with a relative information advantage, as the time horizon increases, this advantage diminishes. Formally, this pattern is due to the asymptotic optimality of the UCB and LOLA algorithms, implying that as the time horizon approaches infinity, the initialization has a negligible effect on the average performance over an infinitely long period. We also see that E\&C consistently underperforms compared to LOLA because it neither leverages LLM predictions nor achieves asymptotic optimality. While E\&C shows better performance as the horizon increases, even when $\tau=1000$, LOLA is still better than E\&C by 5.08\%, which is a large improvement in the media and publishing industry. 

Third, comparing the benchmark algorithms with each other, we see that when the horizon is short, the Pure LLM does better, while UCB is better when the horizon is longer.  In general, this reflects the intuition that active learning is not very useful when the time horizons are very short since the algorithm has no time to explore and learn the CTRs effectively (and then exploit this learning). So simply going with the pure LLM predictions is better. In contrast, when the horizons are longer, UCB (or active learning) outperforms using static predictions from the LLM. Note that LOLA combines the strengths of both UCB and Pure LLM, and hence it is able to outperform both and provide superior performance in both short and long horizons. We also note that E\&C is generally worse than both Pure LLM and UCB, which suggests that the current Upworthy practice is suboptimal. This is because E\&C neither exploits LLM information and nor does it actively learn over time. 

Finally, we will discuss the performance of UCB with LLM priors. Comparing LOLA (2-UCBs) with UCB with LLM priors, the key difference lies in whether the regular UCB $U^1$ is included in the algorithm. The core trade-off here is between the large potential error of $U^2$ when the LLM's predictions are incorrect and the stochastic noise in $U^1$, especially in the early stages. Our observations confirm this trade-off. In short time horizons $\tau = 50$, LOLA underperforms compared to UCB with LLM priors, by a significant margin of $-0.74\%$. This is because the error in the LLM-based CTR predictions is smaller than the stochastic noise of the regular CTR estimator at the start, and including the noisier $U^1$ reduces performance. However, as the time horizon increases (e.g., $\tau = 800$ and $\tau = 1000$), LOLA significantly outperforms UCB with LLM priors by $0.37\%$ and $0.47\%$, respectively. This improvement occurs because, over longer periods, the error in LLM-based CTR predictions becomes more significant than the stochastic noise of the regular estimator, which can also be observed by comparing the performance between the standard UCB and Pure LLM methods. Incorporating the unbiased and less noisy $U^1$ helps enhance performance in these longer horizons. This observation suggests that it is especially beneficial to use 2-UCBs in longer horizons.

In summary, we find that LOLA outperforms existing approaches for experimentation by leveraging the strengths of LLMs and combining them with the benefits of active/online learning. In particular, our results demonstrate the value of LOLA as the time horizon increases, highlighting the limitations of relying solely on pure LLM predictions or the naive E\&C approach.

\subsection{LOLA Variants and Extensions}
\label{ssec:lola_extensions}
So far, we have considered a version of LOLA that uses a fine-tuned CTR prediction model for LLM predictions in the first stage and an LLM-2UCBs algorithm that minimizes regret for online learning in the second step. However, LOLA is a general framework and can be easily adapted to a wide variety of settings. We highlight a few natural extensions and variants below. Details of the models and numerical results (when applicable) are shown in Web Appendix $\S$\ref{appsec:lola_extensions}.

\textbf{Best Arm Identification:} So far, we focused on the goal of regret minimization, where the goal is to maximize the clicks/reward (per Equation \eqref{eq:regret}). While regret minimization is the most commonly studied goal in active learning and the natural goal for businesses in most settings, another commonly studied goal is Best Arm Identification (BAI). The goal of BAI problems is to identify the arm with the highest reward as fast and accurately as possible, regardless of accumulated regret. Formally, given a low failure rate $\delta$, we aim to find the arm $a^*$ with the highest expected reward, at least $1-\delta$ probability, while minimizing the number of total pulls.\footnote{This is also called fixed-confidence setting \citep{garivier2016optimal}. Another common approach is the fixed-budget setting, where the goal is to maximize the probability of finding the best arm within a fixed number of pulls.} We can easily modify the LOLA framework for a BAI goal. In Web Appendix $\S$\ref{appssec:BAI}, we present a LLM-BAI algorithm that builds on the recently proposed BAI algorithm by \citet{mason2020finding}. This algorithm is particularly effective in our setting since it can accommodate scenarios where the difference between the rewards of arms is small or insignificant. We show that our LLM-BAI algorithm significantly outperforms both the plain BAI algorithm (proposed by \citet{mason2020finding}) as well as the standard A/B test under the same failure rate restriction. Please see 
Web Appendix $\S$\ref{appssec:BAI} for a detailed description of the LLM-BAI algorithm and its performance against the standard benchmarks. 

\textbf{Thompson Sampling:} Our bandit specification in $\S$\ref{ssec:bandit_framework} reflects a stochastic bandit setting rather than a Bayesian bandit setting. However, our proposed algorithms can easily be applied in a Bayesian setting. We present a Thompson sampling version of our proposed LOLA algorithm (LLM-TS) and its numerical performance in Web Appendix $\S$\ref{appsec:ts}. We see that UCB-based algorithms perform better than Thompson Sampling-based algorithms in both the LLM-assisted version and the standard version. That is, LLM-TS  performs worse than LLM-2UCBs, and standard TS performs worse than UCB. This pattern has been observed in the literature, and it is generally known that TS suffers from over-exploration \citep{min2020policy}; in other words, UCB's upper confidence shrinks faster, leading it to exploit more effectively. Nevertheless, in business settings where the firm already has a TS-based adaptive experimentation framework, it is easy to adapt the LOLA approach within this approach.


\textbf{User Information:} So far, we have not considered user-level features or how to personalize the content shown to different users based on their behavioral/context features (mostly because the UpWorthy data does not have user-level features). Prior research has shown that using such features can significantly improve the match between users and content \citep{li2010contextual, yoganarasimhan_2020, rafieian_yoganarasimhan_2021}. It is, however, easy to expand our LOLA approach to include user features in both the LLM training phase as well as the online learning phase. We present two versions of LOLA that extend existing contextual bandit algorithms to our setting. First, we can extend the standard contextual linear bandits \citep{chu2011contextual} to our LOLA framework by incorporating both user and text features in the LLM training and online learning phases of LOLA. However, this linear bandit approach has limitations since it assumes that the rewards model is linear in text and user features.  Therefore, we also propose another solution based on the FALCON algorithm that does not make any linearity assumption \citep{simchi2022bypassing}. We present the details of both these contextual LOLA algorithms and their pros and cons in Web Appendix $\S$\ref{appsec:algs}.

\textbf{Alternative rewards:} In the Upworthy setting, the measure of reward is clicks. However, click-based metrics have raised concerns about promoting clickbait headlines, which may reduce user engagement or spread negativity. As a result, the news and media industry is now evolving to focus more on content quality and longer-term user engagement and retention metrics, although clicks still remain an important measure; see \cite{Upworthy2023} for a more detailed discussion. The LOLA framework can be easily adapted to alternative reward measures that align with new business goals, e.g., time spent on the platform during a session, or time spent on the article (and not just clicks on the headline). 

Apart from these extensions, the LOLA approach is quite general and adaptable on other dimensions as well. For instance, it is agnostic to the exact nature of the LLM-based approach used in the first step. Any of the LLM-based approaches discussed in $\S$\ref{sec:pure_llm} can be used in the first step. This is important since the LLM models and fine-tuning approaches continue to advance quickly. Further, LOLA can also be used in conjunction with content/creatives created by LLMs themselves. For instance, one simple way to leverage fine-tuned models is to use them to generate content (instead of using human-generated headlines/content). Indeed, recent studies have shown that promotional content (ads/emails) generated by fine-tuned LLMs tend to outperform traditional personalized ads and human-generated content in effectiveness \citep{kumar2023generative, angelopoulos2024value}. LLM-generated headlines/content can be used as competing arms/treatments within our LOLA framework. Additionally, the LOLA framework can also be easily extended to content recommendation systems used by social media platforms such as X and Facebook. In summary, the plug-and-play nature of our approach can easily accommodate newer advances and developments at all stages of the implementation process.




\section{Conclusion}
\label{sec:conclusion}

In conclusion, in this paper, we examine if and how firms can leverage the LLMs to enhance content experimentation in digital platforms. First, we examine how well LLMs can predict which content would be more appealing to users. We find that while LLMs provide informative predictions, even the best performing fine-tuned LLMs are unable to perfectly predict which content will be more appealing to users, and match the accuracy/regret of experimentation-based approaches. As such, firms cannot fully trust the prediction results solely from LLMs and replace experimentation. 

We introduce LOLA, a novel experimentation framework that combines the predictive power of LLMs with the adaptive efficiency of online learning algorithms to enhance content experimentation. By leveraging LLM-based prediction models as priors, our approach minimizes regret in real time, leading to significant improvements in total reward. Comprehensive numerical experiments based on a large-scale A/B testing dataset demonstrate that LOLA outperforms the traditional A/B tests and pure online learning algorithms. 

From a managerial perspective, LOLA offers a versatile solution to a broad set of scenarios where the firm needs to decide which content to display to users. LOLA is particularly valuable under limited traffic conditions that are common in many digital media platforms. This includes cases where content becomes stale quickly (e.g., news articles) or social media platforms like X and TikTok, where users generate a lot of new content regularly, and as a result, the amount of content is high relative to the amount of traffic. While our focal application is in the context of headline selection in the news and media industry, the framework can easily be applied to other problems, such as digital advertising, email marketing, and promotions. Indeed, a large literature in marketing focuses on how to use experiments in conjunction with machine learning methods to identify optimal treatments and/or personalize treatments; see \citet{rafieian_yoganarasimhan_2023} for a detailed review. The LOLA approach can be easily used in many of those settings and help with maximizing outcomes of interest while simultaneously reducing the cost of experimentation. Finally, from an implementation perspective, LOLA offers many advantages. Since it can utilize open-source LLM models, it is cost-effective and can be easily deployed across different tasks once LLM models are fine-tuned on relevant datasets. The fine-tuning process itself is cost-efficient and can be performed on entry-level GPUs, making advanced LLM-based techniques accessible to firms with limited budgets. Relying on open-source LLMs also ensures that the firm's data stays within the firm and is not accessible to the owners of proprietary LLM models (such as Google and OpenAI). Further, our approach is compatible with many bandit-based experimental systems used by large digital firms such as Amazon \citep{fiez_etal_2024}, and as such, is easy to integrate and deploy.



\section*{Code and Data}
The code used in this study is publicly available at \href{https://anonymous.4open.science/r/LOLA_LLM_Assisted_Online_Learning_Algorithm_for_Content_Experiments}{LLM News Github Repository} to enable the reproduction of our results and to support future research that applies LLMs to experimentation and business problems. 


\section*{Funding and Competing Interests Declaration}
Author(s) have no competing interests to declare.


\end{bibunit}

\newpage
\setcounter{table}{0}
\setcounter{figure}{0}
\setcounter{equation}{0}
\setcounter{page}{1}
\renewcommand{\thetable}{A\arabic{table}}
\renewcommand{\thefigure}{A\arabic{figure}}
\renewcommand{\theequation}{A\arabic{equation}}
\renewcommand{\thepage}{\roman{page}}
\singlespacing
\renewcommand{\thesection}{\Alph{section}}
\pagenumbering{roman}

\begin{bibunit}

\begin{appendices}

\section{User Survey}
\label{appsec:survey}

We conduct a survey of undergraduate students from a large public university. The goal of the survey was to examine whether lay human users can accurately identify which headlines would lead to more clicks. Recall that the headlines themselves were created by Upworthy editors, but those editors need an experiment to identify the most appealing headline.

We design two similar surveys using Qualtrics. The first survey randomly samples 3,000 pairs of headlines from 39,158 headline-pairs that are significantly different from each other (at a 0.05 significance level), along with 500 pairs that have no significant differences in CTRs. The second survey utilizes the same 3,000 significant samples as the first but does not include the insignificant samples. 

In both surveys, students were asked to select the catchier headline from a pair of headlines. The introductory text for the surveys was:\\
``Please help us understand what makes a news headline appealing to readers. In this survey, you will be presented with 20 pairs of headlines. Please choose the one that captures your attention more effectively or feels more engaging in each pair, assuming both headlines relate to the same news content. There are no right or wrong answers; we are simply interested in your opinions.'' \\
Respondents then proceed to a new page featuring 20 headline comparison questions with the prompt, ``Please select the headline that you find more attractive,'' with only one permissible response. In all the comparisons, we randomly swap the order of the headlines to ensure that the display order does not correlate with the correct answer. We also incorporated two attention checks in the second survey.

The first survey included an additional response option: ``Both headlines seem equally good.'' At the survey's conclusion, we also collected data on users' news consumption habits. Specifically, we ask, ``How often do you read the news?'' with options ``Never'', ``1-2 days per week'', ``3-4 days per week'', ``5-6 days per week'', ``Everyday''.


Detailed information and survey results are documented in Table \ref{tab:survey}. Some students did not complete the survey or skipped some comparisons, resulting in 3,809 responses out of a potential 4,040 ($202\times20$). To calculate the accuracy metric, we focused solely on the significant pairs, and in the first survey, we excluded the uncertain answer ``Both headlines seem equally good'' to facilitate a fair comparison between the two surveys. Accuracy here aligns with the LLM tests. We regarded the headline with the higher CTR as the correct answer and calculated the proportion of correct responses as accuracy. As such, we should expect a 50\% accuracy rate from random guessing in this paired comparison.

The survey results indicate that respondents' ability to identify catchier headlines is comparable to random guessing -- the accuracy rate in both surveys is not significantly different from 50\% at the 0.05 significance level. Furthermore, accuracies did not vary significantly across different groups based on news reading frequency. These findings suggest that the human ability to discern catchier headlines is limited, highlighting the intrinsic difficulty of the task.

\begin{table}[ht]\centering
\begin{tabular}{llccc}\toprule
& &Survey 1 &Survey 2 \\
\midrule
\multicolumn{2}{l}{conducted time} &4/14 - 4/21 &4/29 - 5/17 \\
\multicolumn{2}{l}{\# of respondents} &202 &339 \\
\multicolumn{2}{l}{headline pair pool} &3000 significant + 500 insignificant pairs &3000 significant pairs \\
\multicolumn{2}{l}{\# of sampled pairs in survey} &20 &20 \\
\multicolumn{2}{l}{\# total received responses} &3,809 &4,571 \\
\midrule
\multicolumn{2}{l}{overall accuracy \%} &45.22 &51.58 \\
\midrule
 &\multicolumn{3}{c}{Results based on different answer to ``How often do you read the news?''} \\
\cline{2-4}
\multirow{2}{*}{Never} &proportion \% &21.78 &16.22 \\
&accuracy \% &45.76 &51.62 \\
\cline{2-4}
\multirow{2}{*}{1-2 days per week} &proportion \% &48.51 &49.56 \\
&accuracy \% &45.36 &52.54 \\
\cline{2-4}
\multirow{2}{*}{3-4 days per week} &proportion \% &18.81 &18.29 \\
&accuracy \% &46.83 &51.39 \\
\cline{2-4}
\multirow{2}{*}{5-6 days per week} &proportion \% &4.95 &5.31 \\
&accuracy \% &39.05 &50.15 \\
\cline{2-4}
\multirow{2}{*}{Everyday} &proportion \% &5.94 &8.26 \\
&accuracy \% &42.35 &47.48 \\
\bottomrule
\end{tabular}
\caption{Survey Response Analysis. No accuracies in this table are significantly lower or higher than 50\% at 0.05 significant level. 8 respondents did not finish Survey 2 in the limited given time, so the summation of proportion from different subgroups is smaller than 100\% in Survey 2.}\label{tab:survey}
\end{table}

\section{Appendix for Prompt-based Approaches}

\subsection{Pairwise Comparison Between GPT Prompts}
\label{appsec:pair_t}

\begin{figure}[ht]
    \centering
    \includegraphics[width=0.85\linewidth]{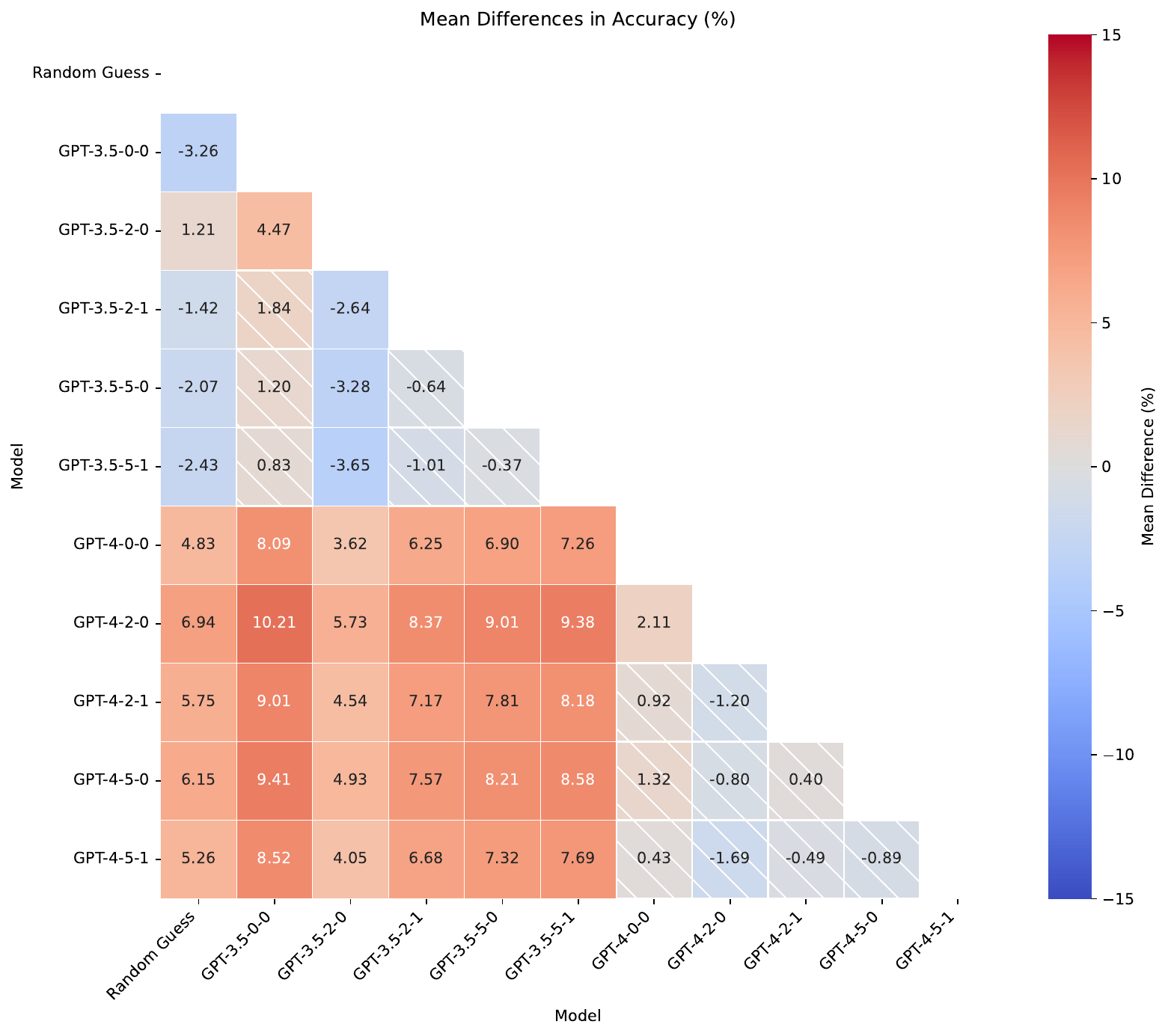}
    \caption{Average accuracy differences between different GPT prompts. The number in the heatmap presents the accuracy of prompts indicated on the left minus the accuracy of prompts indicated at the bottom. Shaded blocks indicate insignificant differences.}
    \label{fig:gpt_ttest_mean}
\end{figure}

\begin{figure}[ht]
    \centering
    \includegraphics[width=0.85\linewidth]{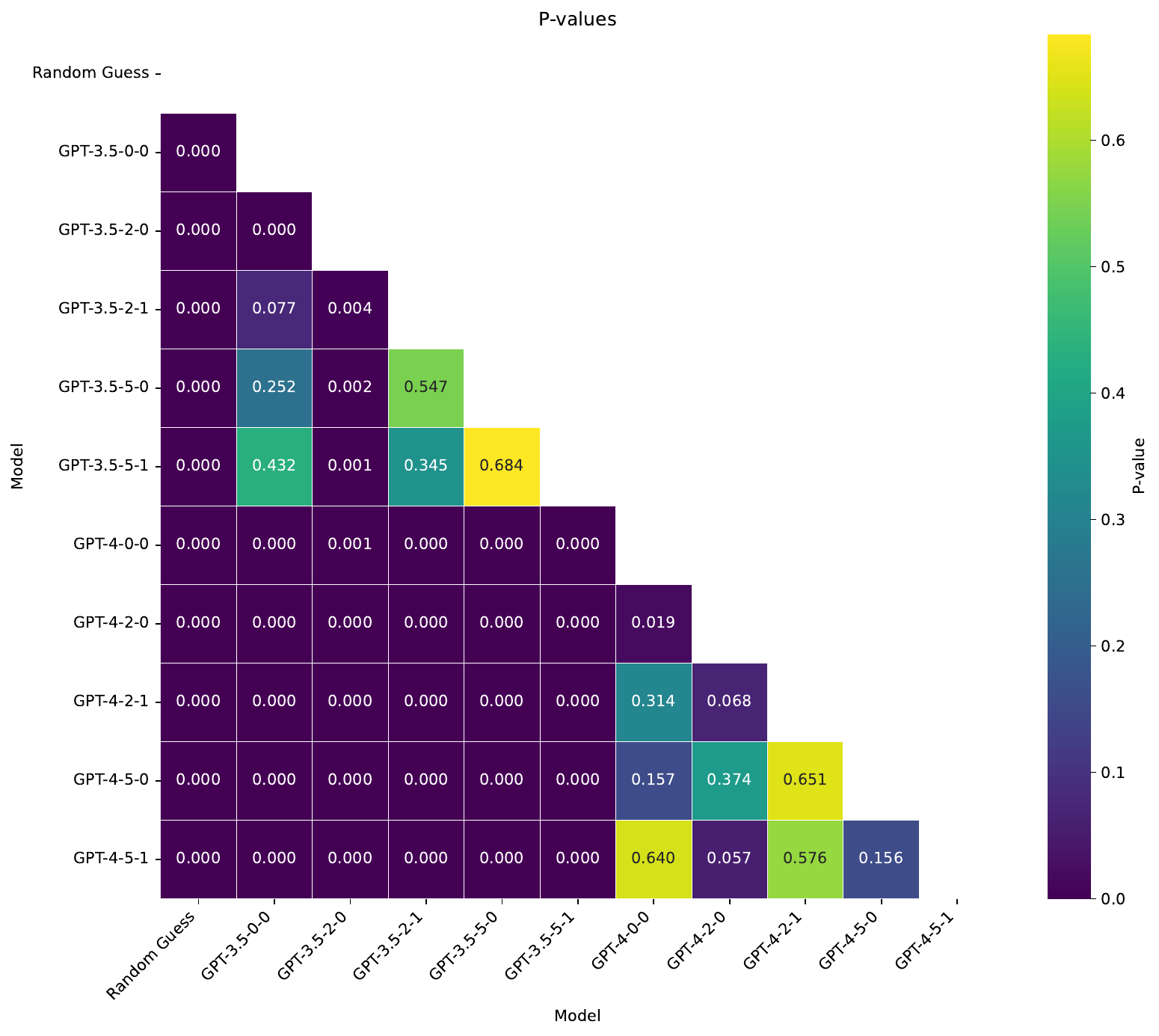}
    \caption{The number in the heatmap presents the p-value from the paired t-test between different GPT prompts.}
    \label{fig:gpt_ttest_p}
\end{figure}

To determine whether one prompt is significantly better than the other, we record a binary indicator of the correct/incorrect answer for each prompt on each sample and then conduct a paired t-test on correct/incorrect answer indicators to see whether the answers from one prompt are significantly more accurate than the other. The mean differences and p values are documented in Figure \ref{fig:gpt_ttest_mean} and Figure \ref{fig:gpt_ttest_p}. The nomenclature for GPT prompts in these two figures follows a specific format. The first component is the name of the GPT model, the second component indicates the number of demonstrations, and the third component specifies whether the outputs in the demonstrations are correct or not. A value of 0 indicates the correct label, while a value of 1 indicates the wrong label. For example, GPT-4-0-0 means zero-shot prompting using GPT-4, and GPT-3.5-5-1 means in-context learning with 5 demonstrations and wrong labels using GPT-3.5.

\section{Appendix to Embedding-based Approaches}

\subsection{Implementation Details}
\label{appsec:MLP fine tune}

We start by introducing three embedding models used:
\squishlist
\item \textbf{OpenAI embeddings:} At the time of this study, the latest and best embedding model available from OpenAI was \texttt{text-embedding-3-large}. These embeddings are obtained using the \href{https://platform.openai.com/docs/guides/embeddings}{OpenAI embedding API} in Python. This embedding model generates embeddings with up to 3,072 dimensions and supports a maximum token limit of 8,191, with its knowledge base extending up to September 2021. A notable feature of this model is its flexibility -- it allows users to tailor the embedding dimension to their specific needs.\footnote{Other models capable of producing text embeddings, like BERT \citep{devlin2018bert} and Sentence-BERT \citep{reimers2019sentence}, have fixed embedding vector dimensions that remain constant once the model is trained. These dimensions cannot be changed or resized for different applications. However, OpenAI's embedding model uses a technique that, while producing outputs of fixed length, encodes information at different granularities \citep{kusupati2022matryoshka}. This allows a single embedding to adapt to the computational constraints of downstream tasks. More specifically, downstream tasks can use only the first several dimensions of the original vector, and the information retained decreases smoothly as a shorter part of the vector is used.} In our experiments, we consider both 3,072- and 256-dimensional embeddings to examine whether the size of the embedding vector size affects the performance of the downstream CTR prediction models.\footnote{A higher dimension does not necessarily mean better performance. While a higher dimension can retain more information for each sample, it also means that the samples are spread more sparsely in a higher-dimensional space given a limited number of training samples. This sparsity can increase the possibility of over-fitting.} 

\item \textbf{Llama-3-8b:} Meta's Llama-3 \citep{metallama3} is regarded as one of the most advanced open-source LLMs to date \citep{llama3modelcard}. We obtain embeddings from Llama-3 with 8 billion parameters (Llama-3-8b). We fine-tune this exact model using LoRA in $\S$\ref{ssec:lora}; therefore, using the embeddings from this model allows for a direct comparison of fine-tuning vs. embedding-based methods (see $\S$\ref{ssec:llm_summary} for comparisons). Llama-3-8b does not produce embeddings explicitly designed to represent the semantic meaning of entire input sequences. To address this issue, we apply the technique described in \cite{behnamghader2024llm2vec}, which transforms the Llama-3-8b model into a more suitable text embedding model, outputting embeddings with a dimension of 4,096. 

\item \textbf{Word2Vec embeddings:} 
Word2Vec, proposed by \citet{mikolov_etal_2013a}, was the leading embedding model before the widespread adoption of Transformers in the Natural Language Processing (NLP) literature. Unlike LLM-based text embeddings, Word2Vec does not use Transformers and cannot pay attention to surrounding token information. Embeddings from Word2Vec have been used in the recent marketing literature in a variety of use cases, including product attribute modeling \citep{timoshenko_hauser_2019, wang_etal_2022}, the effect of creativity on product popularity \citep{sozuer_etal_2024}. We compute the Word2Vec embedding vector for each word in the headline and average them to get the headline embedding, a common practice in Word2Vec. To ensure consistency with the dimensions used for OpenAI embeddings, we consider Word2Vec embeddings with 256 and 3,072 dimensions. 

\item \textbf{BERT embeddings:} BERT (Bidirectional Encoder Representations from Transformers), introduced by \citet{devlin2018bert}, is a seminal model in natural language processing that generates context-sensitive embeddings for text. Unlike earlier models that produce static word embeddings, BERT captures the nuances of word meanings based on their context within a sentence, resulting in more accurate representations. The standard BERT model produces embeddings with 768 dimensions, which are fixed once the model is trained. This fixed dimensionality can be a limitation when adapting to tasks with varying computational constraints. Despite this, BERT’s ability to understand context has led to its widespread adoption in various NLP applications, including sentiment analysis, question answering, and named entity recognition.
\squishend

\noindent To assess the impact of model complexity on performance, we consider using a simple linear model and a multilayer perceptron (MLP) model. MLP is a powerful neural network architecture consisting of multiple linear and non-linear hidden layers that allow the model to learn complex patterns in the data. We use a fully connected neural network with ReLU activation for hidden layers, with dropout layers added to prevent over-fitting. We refer interested readers to Chapter 5 in \cite{zhang2021dive} for more details. We train the MLP using the stochastic gradient descent algorithm to minimize binary cross-entropy and perform hyper-parameter tuning with Bayesian Optimization to determine the optimal configuration, with the training process and the architecture details below.

\textit{Hyperparameters of the network structure.} We employ a fully connected neural network with ReLU activation functions for the hidden layers, a configuration widely used in practice. The network consists of two hidden layers, each with 128 nodes. To mitigate overfitting, we incorporate dropout after each hidden layer. Dropout layers randomly deactivate neurons based on a specified dropout rate while scaling up the remaining neurons to maintain the same mean of the layer. We set the dropout rate to 0.5. We did not further test other hyperparameter settings, as this configuration already delivers sufficiently good performance. However, more careful hyperparameter optimization could potentially improve results.

\textit{Training Procedure Hyperparameters.} The MLP is implemented in PyTorch, with the data split into 70\% for training, 10\% for validation, and 20\% for testing, as introduced in $\S$\ref{sec:pure_llm} . To prevent the network from memorizing the order of samples, we shuffle the training set at the start of each epoch. Determining the appropriate batch size is crucial, as large batch sizes can lead to convergence to local optima, while small batch sizes may slow down convergence. We balance accuracy and training speed by setting the batch size to 10 and the learning rate to $0.01$. The loss is calculated as the mean squared error between the real CTR and the predicted CTR, and we use stochastic gradient descent for training.

\subsection{Performance Comparison of Different Embeddings}
\label{appssec:embed_all}
\begin{table}[htp!]
\centering
\begin{tabular}{lll}
    \toprule
    {Embedding Model} & {CTR Prediction}& {Accuracy on the test set}\\
    \midrule
    \multirow{2}{*}{OpenAI-256E} & \centering Linear & \textbf{46.28\%} \\
                                 & \centering MLP &  44.38\% \\
    \midrule
    \multirow{2}{*}{OpenAI-3072E} & \centering Linear & 43.06\% \\
                                  & \centering MLP &  45.17\%  \\
    \midrule
    \multirow{2}{*}{Llama-3-8b-4096E} & \centering Linear & 42.78\%  \\
                                     & \centering MLP   & 36.03\%  \\     
    \midrule
    \multirow{2}{*}{Word2Vec-256E} & \centering Linear & 41.07\%  \\
                                 & \centering MLP   & 36.96\%  \\     
    \midrule
    \multirow{2}{*}{Word2Vec-3072E} & \centering Linear & 40.27\% \\
                                & \centering MLP & 37.27\%  \\
    \midrule
    \multirow{2}{*}{BERT} & \centering Linear & 42.32\% \\
                                & \centering MLP & 35.89\%  \\
    \bottomrule
\end{tabular}
\caption{The performance of different embedding and CTR prediction models on test data. All models use the same training data, and accuracy is evaluated on the test data. Note that the Linear model's training is a deterministic process after fixing the data, while MLP training incurs randomness via stochastic gradient descents. Therefore, we rerun MLP training three times with different random seeds and report the average accuracy from three reruns.
}
\label{tab:embedding_classification_results_all}
\end{table}

The comparison of different embeddings and CTR prediction models are listed in Table~\ref{tab:embedding_classification_results_all}. We observe that OpenAI embeddings outperform Llama-3-8b, Word2Vec, and BERT-based models. This is because OpenAI's LLMs are state-of-the-art models that naturally perform better than the other embeddings. Additionally, the performance of embeddings from the open-source Llama-3-8b falls between that of the proprietary OpenAI model and Word2Vec/BERT, which does not use the transformer architecture.

\subsection{Effect of Training and Test Data Size on Performance}
\label{appssec:sample_size_expts}

In this analysis, we focus on OpenAI-256E (as it provides the best performance) and use linear regression as our prediction model, since the MLP does not significantly outperform the linear model on OpenAI embeddings and is easier to train.

\begin{figure}[ht]
\centering
\includegraphics[width=0.8\textwidth]{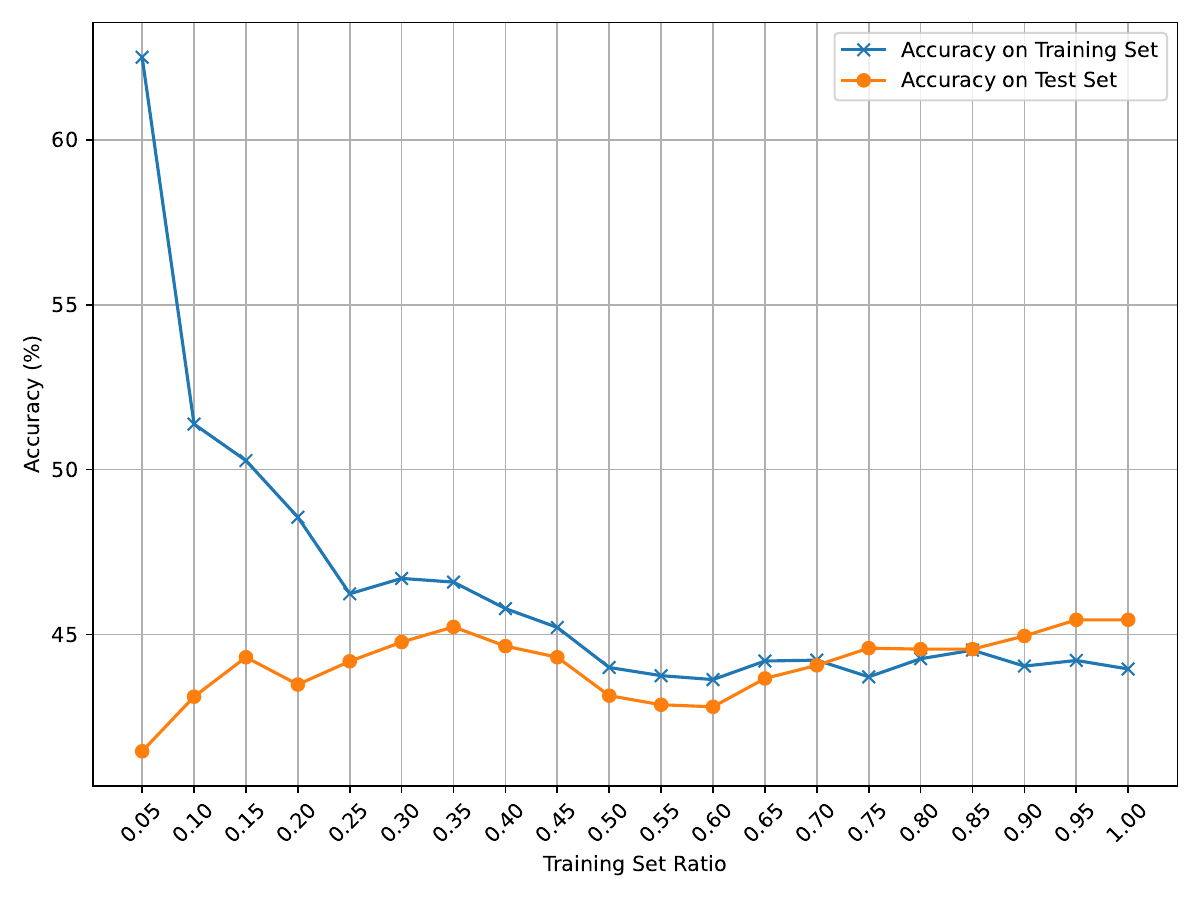}
\caption{Performance of CTR prediction model with OpenAI-256E under varying training dataset sizes}
\label{fig:Results_proportion_ratio_regression}
\end{figure}

To assess the effect of training sample size on accuracy, we vary the training data size and measure performance. Specifically, we set a training size ratio, randomly select a subset of articles based on this ratio from the full training set, and use this subset to train the model. As shown in Figure \ref{fig:Results_proportion_ratio_regression}, increasing the amount of training data leads to a gradual improvement in test accuracy. For example, increasing the training size from 20\% of the total dataset (about 30\% of the training set) to 70\% of the total dataset (100\% of the training set) raises accuracy from 44\% to 46\% on the test data. This observation also indicates that, achieving a significant performance boost in LLMs would require a much larger dataset than is currently available.

\subsection{Potential Threats and Robustness Checks}
\label{appssec:threats}

Note that in the following two analyses of threats, we still focus on OpenAI-256E with linear regression as our prediction model as it provides the best performance.

Next, we discuss two potential threats to the inference from our analysis here and our robustness checks to alleviate concerns around these threats. First, one may be concerned about whether Upworthy's data has been used as part of OpenAI's training corpus, which could potentially invalidate our analysis. However, this is unlikely for several reasons. First, it is improbable that OpenAI would have specifically downloaded these Upworthy CSV files (Comma-Separated Values files) dataset from a subfolder in \href{https://osf.io/jd64p/}{Upworthy Research Archive}. Second, the original data only contains impressions and clicks for each headline, as shown in Table \ref{tab:data in upworthy}, rather than CTR information. Therefore, during the training of OpenAI models, it is unlikely they were exposed to the CTR information since the training objective is to predict the next word. Third, we further conducted an experiment as the robustness check to determine if this dataset was indeed a part of OpenAI's corpus. We used embeddings from two headlines with significantly different impressions (possibly from different tests) for the downstream task of predicting which headline received higher impressions. If the embedding model was trained on this corpus, its embeddings should reflect the number of impressions and perform well on this task. However, the observed accuracy was approximately 50\%, indicating that the embeddings do not even contain such impression information, confirming our assumption. 

Second, one may be concerned that readers' preferences changed over time and that the current sample splitting method fails to take into account when the headlines were created, potentially making our CTR prediction model invalid if users' preferences shift. To address this concern, we conducted a robustness check by splitting the dataset based on the creation time of the headlines. Specifically, we used the initial 70\% of the data, in chronological order, as the training set, and the latest 20\% as the test set (the remaining 10\% was reserved for algorithm hyperparameter fine-tuning). We found that the accuracy was 46.25\% on the test data, which is close to the embedding performance on the randomly split training and test dataset, with an accuracy of 46.28\%, as given in Table \ref{tab:embedding_classification_results}. This indicates that time-based preference shifts are not a concern in our data.

\section{Appendix to LoRA Fine-Tuning}
\label{appsec:app_lora}
\subsection{Transformer and Llama Architecture}
\label{appsec:llama_transformer}
The Transformer model, introduced by \cite{vaswani2017attention}, has revolutionized the field of NLP with its innovative attention mechanism and highly parallelizable architecture. Unlike traditional recurrent neural network (RNN) based models, the Transformer uses self-attention mechanisms to process input data. This shift allows for greater computational efficiency and significantly improved performance across various NLP tasks.

\begin{figure}[ht]
    \centering
    \includegraphics[width=0.9\linewidth]{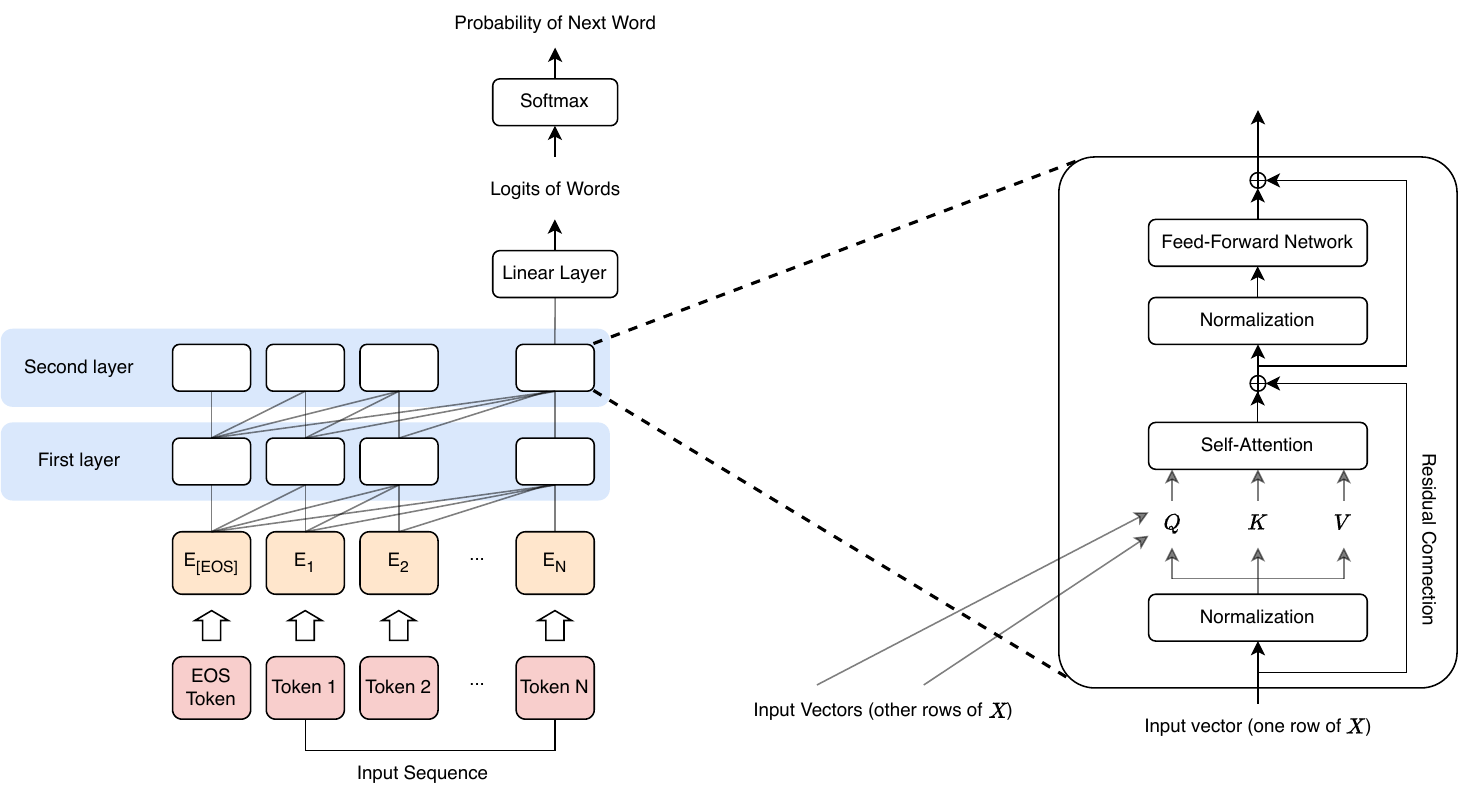}
\caption{Illustration of the Transformer block and architecture typical in most LLMs. \\
The left portion of the figure displays the overall architecture of an LLM, highlighting the layers of Transformer blocks and the connections that represent the attention mask. Lines between Transformer blocks in different layers indicate where attention is applied; the absence of a line signifies no attention interaction. Specifically, a causal attention mask is depicted, ensuring that predictions for each token depend only on preceding tokens. Note that this figure represents one type of attention mask; other types are also employed in various LLMs but are not depicted here. Additionally, only two layers are shown for simplicity; actual models may contain many more layers and connections, which are not depicted here to maintain clarity and focus on key components. \\
The right portion provides a detailed view of a single Transformer block, including internal components. We omit the depiction of multi-head attention for brevity. Note that different LLMs may use slightly different designs.}
    \label{fig_transformer}
\end{figure}

As illustrated in the right portion of Figure \ref{fig_transformer}, the Transformer block comprises multiple layers, each featuring two primary components: a multi-head self-attention mechanism and a position-wise fully connected feed-forward network. Key enhancements include layer normalization \citep{ba2016layer} and the use of residual connections \citep{he2016deep}, which improve training stability and help prevent issues such as vanishing gradients. While some specifics are omitted to maintain brevity and readability, it's important to note that such architectures are subject to rapid evolution and vary across different LLMs.

\textbf{Self-Attention Mechanism:} The most important innovation in the Transformer block is the self-attention mechanism. It enables each token in a sentence to ``pay attention'' to all other words when computing its own representation. This is achieved using an attention mask that guides which tokens should influence each other, thereby allowing the model to capture dependencies between words regardless of their distance in the text. Specifically, for a given input sentence, the model first converts each token into an embedding,\footnote{This embedding differs from the text embedding used in the previous section. Each embedding here simply corresponds to one token without paying attention to the whole input sequence. The text embeddings used in the previous section are essentially the contextualized embeddings, which are generated by passing the tokens through multiple layers of Transformers, capturing rich contextual information, and summarizing the whole input sequence.} forming a matrix \(X\) where each row corresponds to the embedding of a token. If the sentence contains \(n\) tokens, and each token is represented by a \(d\)-dimensional embedding, then \(X\) is an \(n \times d\) matrix. From these embeddings, the model generates three matrices: Query \(Q\), Key \(K\), and Value \(V\):
\[ Q = XW_Q, \quad K = XW_K, \quad V = XW_V \]
where \(W_Q\), \(W_K\), and \(W_V\) are trainable weight matrices of dimensions \(d \times d_q\), \(d \times d_k\), and \(d \times d_v\) respectively. \(d_q\) and \(d_k\)  are typically set to be the same, allowing for the dot-product operation used in computing the attention scores to be valid.

The attention scores are computed by first taking the dot product of the query matrix \(Q\) and key matrix \(K\), then scaling by the square root of the key vector dimension \(d_k\), and applying an attention mask \(M\). A softmax function is applied subsequently:
\[ \text{Attention}(Q, K, V) = \text{Softmax}\left(\frac{QK^\top}{\sqrt{d_k}} + M\right)V \]
The attention mask \(M\) typically includes negative infinity at positions where attention is not applicable, ensuring the attention weights are effectively zero after applying the softmax. At positions where attention is relevant, the mask values are zero. In models like GPT and Llama, a causal attention mask, a.k.a, autoregressive attention, is utilized to prevent future tokens from influencing the prediction of the current token during training. The causal mask is a lower triangular matrix, where positions above the diagonal are set to negative infinity, and positions on and below the diagonal are set to zero. While BERT employs a bidirectional attention mechanism that allows each token to depend on all other tokens in the sequence.

\textbf{Multi-Head Attention:} To capture diverse aspects of the relationships between tokens, the Transformer enhances the self-attention mechanism through the use of multi-head attention. This approach involves performing multiple self-attention operations in parallel, with each ``head" having its own independent \(Q\), \(K\), and \(V\) matrices. The outputs from these operations are concatenated and then linearly transformed to produce the final attention output:
\[ \text{MultiHead}(Q, K, V) = \text{Concat}(\text{head}_1, \ldots, \text{head}_h)W_O, \]
where each head computes an independent self-attention process:
\[ \text{head}_i = \text{Attention}(XW_Q^i, XW_K^i, XW_V^i), \]
with \(W_Q^i\), \(W_K^i\), and \(W_V^i\) being the learnable parameter matrices specific to the \(i\)-th head. \(W_O\) is the weight matrix with dimension \(h d_v\times d\), and is applied to the concatenated outputs from all heads, integrating the distinct perspectives captured by each head into a unified output.

\textbf{Feed-Forward Networks:} Each layer of the Transformer also includes a position-wise fully connected feed-forward network. This network comprises two linear transformations with a ReLU activation function in between:
\[ \text{FFN}(x) = \max(0, xW_1 + b_1)W_2 + b_2 \]
where \(W_1\), \(W_2\), \(b_1\), and \(b_2\) are learnable parameters. The final output of the FNN has the same dimension \(d\) as the input embeddings. This network acts on each position separately and identically, providing additional non-linearity and enhancing the model's ability to learn complex patterns in the data.

\textbf{Llama Architecture:} The Llama model enhances the basic Transformer architecture to handle vast amounts of data and leverage extensive computational resources effectively, establishing it as one of the most powerful open-source language models currently available. Llama-3 retains essential components such as multi-head self-attention and feed-forward layers from the original Transformer model but incorporates several techniques to stabilize training and boost efficiency and performance. These enhancements include the application of pre-normalization with RMSNorm \citep{zhang2019root}, the SwiGLU activation function \citep{shazeer2020glu}, rotary positional embeddings \citep{su2024roformer}, and grouped-query attention \citep{ainslie2023gqa}, among others. Like Llama-3, other LLMs share a similar Transformer architecture but vary in the number of Transformer layers, training corpora, size of the weight matrices, activation functions, etc.

In the multi-layer structure of LLMs, the output of one layer serves as the input to the subsequent layer. Each layer's output is a matrix of the same dimensions as the input matrix \(X\), with each row representing the transformed embedding of a token, enriched with context from other tokens through the self-attention mechanism. After processing through all the layers, the final output matrix contains the contextualized embeddings of all tokens in the input sentence. Unlike RNNs that process input sequences one token at a time, the Transformer architecture can process whole input sequences in parallel, significantly enhancing processing speed and efficiency.

It is worth noting that LLMs do not necessarily operate in a conversational manner, nor do they always use language as their output. As we will demonstrate in the following, we employ the Llama model as a CTR prediction model. In applications like conversational AI, exemplified by GPT, the contextualized embeddings from the last token of the last layer of the Transformer are fed into a linear layer followed by a softmax function, the configuration of which transforms these embeddings into a probability distribution over the vocabulary, enabling the model to predict the next token in the sequence.

\subsection{LoRA Fine-Tuning Technique}
\label{appsec:lora_intro}
\begin{figure}[ht]
    \centering
    \includegraphics[width=0.8\linewidth]{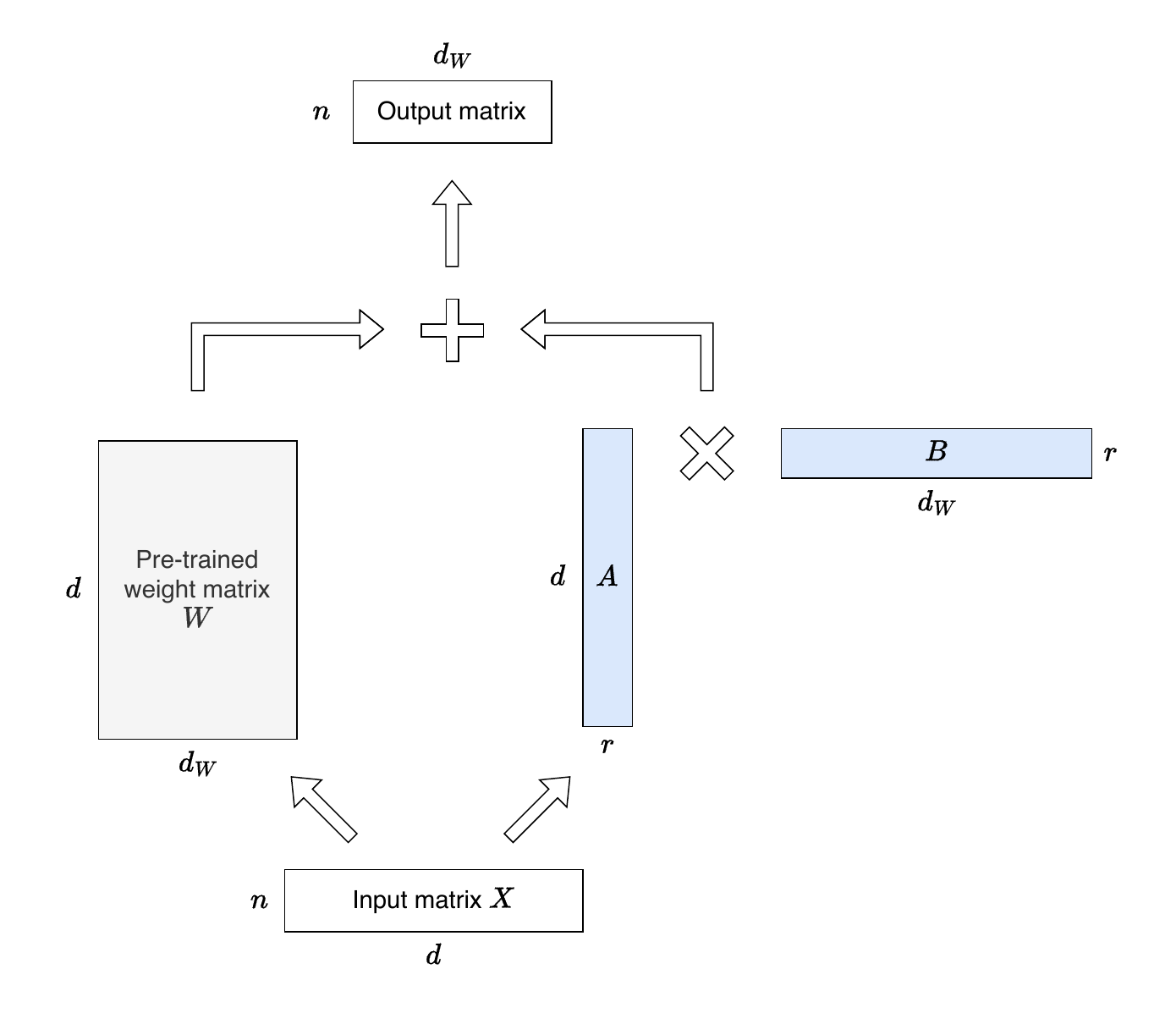}
\caption{Illustration of LoRA Fine-Tuning.}
    \label{fig_lora}
\end{figure}

We illustrate LoRA in Figure \ref{fig_lora}. We use $W$ to represent any pre-trained weight matrix, which can be either $W_Q$, $W_V$, or $W_K$. Specifically, $W_Q$ is the weight matrix for the query vectors, $W_K$ is for the key vectors, and $W_V$ is for the value vectors in the attention mechanism of the Transformer architecture. These matrices store most of the information and knowledge learned by LLM, and are crucial for calculating the attention scores and subsequently determining the relevance of different tokens in the input sequence. For more technical background, we refer to Web Appendix $\S$\ref{appsec:llama_transformer}. We use $d \times d_W$ to represent the dimension of the weight matrix and $r$ to represent the rank of the low-rank matrices. We denote the modified weight matrices after fine-tuning as $W'$, which can be further expressed as:
\[
W' = W + \Delta W, \quad \mathrm{with} \quad \Delta W = A B, \quad \mathrm{where} \quad A \in \mathbb{R}^{d \times r}, \quad B \in \mathbb{R}^{r \times d_W}.
\]

LoRA essentially introduces the low-rank matrices $A$ and $B$ to approximate the weight matrix change $\Delta W$ after the fine-tuning. These low-rank matrices are updated during the fine-tuning process, while the original weight matrix $W$ remains frozen. This strategy significantly reduces the number of trainable parameters from $d \times d_W$ to $r \times(d + d_W)$ because the rank $r$ is typically set to a small value, such as 4 or 8, to balance the trade-off between model capacity and computational efficiency.

\subsection{Implementation Details}
\label{appsec:finetune_details}

\begin{figure}[ht]
    \centering    \includegraphics[scale=0.8]{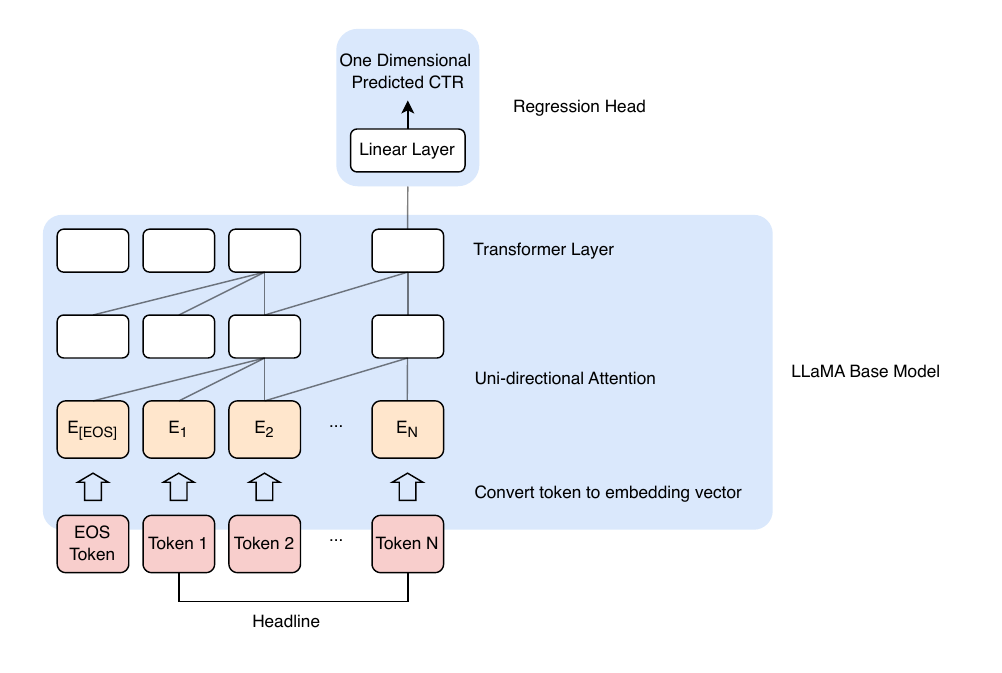}
    \caption{The regression pipeline using the Llama model. Headlines are first tokenized and each token is converted into an embedding vector. For illustrative purposes, only two Transformer layers are shown, although the actual model comprises 32 layers for the 8 billion parameter configuration. The figure highlights the use of uni-directional attention in Llama, whereby each token can only incorporate information from preceding tokens. Connections between Transformer blocks represent the attention mechanism, and only a subset of these links are displayed for simplicity. The output from the last token is fed into a regression head, whose output represents the predicted CTR.}
    \label{fig:llama_frame}
\end{figure}

We use exactly the same training and test sets as in $\S$\ref{ssec:embeddings}, but the inputs here are the original headlines with the text information. We add a regression head to the original Llama-3-8b base model, which includes 32 Transformer layers, as depicted in Figure \ref{fig:llama_frame}. Llama-3-8b employs a causal attention mask, where the attention is unidirectional—each token can only attend to preceding tokens in the sequence. Thus, the last token in the sequence, being informed by all prior tokens, is used for predicting the CTR. This embedding vector of the last token (which is 4096-dimensional), output by the last Transformer layer, is processed through a linear layer producing a one-dimensional scalar that represents the predicted CTR.

During fine-tuning, our objective is to minimize the mean square error on the training set. We implement LoRA fine-tuning using the PEFT library developed by \href{https://huggingface.co/docs/peft/}{Hugging Face}. We set the dimension of the low-rank matrices as $r = 4$. 
Although LoRA can be applied to any subset of weight matrices in LLMs, applying LoRA to query weight matrices $W_Q$ and value weight matrices $W_V$ is a common practice and typically gives better performance \citep{hu2021lora}. Following this practice, in our experiment, we specifically apply LoRA to the $W_Q\in\mathbb{R}^{d\times d_q}$ and $W_V\in\mathbb{R}^{d\times d_v}$ matrices within the Transformer blocks of Llama-3-8b, where $d=4096, d_q=4096, d_v=1024$. To mitigate over-fitting issues, we also apply a relatively large dropout rate of 0.3 to these low-rank matrices.\footnote{Furthermore, we enable 16-bit mixed precision training. Although 16-bit training slightly reduces numerical precision, it accelerates operations on GPUs, which enhances computational efficiency significantly. This mode makes the training process faster and consumes less memory compared to the default 32-bit training, especially for large-scale models like Llama-3-8b.}

With this LoRA configuration on Llama-3-8b, the total number of trainable parameters in the model is 1,703,936, which translates to only 0.023\% of the total 7,506,636,800 parameters. We fine-tune the model for 10 epochs, meaning each sample (pair of headlines) in the training set is used 10 times in the training. A linearly decreasing learning rate scheduler is employed, starting from an initial rate of 2e-5 and decreasing linearly to zero by the end of the fine-tuning process. This strategy helps stabilize the training by gradually reducing the rate of updates to the weights, thus smoothing the convergence and reducing the risk of overshooting minima in the loss landscape.

\subsection{Performance on Different Train-test Splitting}
\label{appsec:split}
Recall that we observe the accuracy on the test set being slightly higher than the accuracy on the training set in Figure \ref{fig:lora_result}. This phenomenon is not specific to the LoRA fine-tuning method; the same effect is observed with embeddings. It is likely due to the training-test split resulting in unbalanced samples in different sets. To investigate this issue, we re-run experiments on different splits and report the training and test performance to validate this explanation. To save computation costs, instead of rerunning LoRA fine-tuning on different splits, we use embeddings for this sensitivity check, specifically the OpenAI-256E embedding with the Linear CTR prediction model, as used previously in $\S$\ref{ssec:embeddings}. The split used in the paper corresponds to random seed 42. We reported the results using seeds 1, 2, 3, 4, and 5 in Table \ref{tab:accuracy_results}. The accuracy on the training and test sets is similar across these splits.

\begin{table}[htp!]
\centering
\begin{tabular}{lccccc}
    \toprule
    & \textbf{Seed 1} & \textbf{Seed 2} & \textbf{Seed 3} & \textbf{Seed 4} & \textbf{Seed 5} \\
    \midrule
    \textbf{Training Set Accuracy} & 39.98\% & 40.51\% & 40.64\% & 40.50\% & 40.24\% \\
    \textbf{Test Set Accuracy}     & 40.85\% & 40.85\% & 39.84\% & 40.91\% & 40.33\% \\
    \bottomrule
\end{tabular}
\caption{Training and test set accuracy for different random seeds.}
\label{tab:accuracy_results}
\end{table}

\section{Cost Comparison of Pure-LLM-Based Methods}
\label{appsec:costs}

In Table \ref{table:cost_effort}, we summarize the monetary cost and effort required for each pure-LLM method used in this paper. 

\begin{table}[ht]
    \centering
    \begin{tabular}{m{1.5cm} m{6cm} m{5.5cm} m{1.5cm}}
    \toprule
    \textbf{Method} & \textbf{Cost} & \textbf{Runing time} & \textbf{Effort} \\
    \midrule
    Prompt & \$15.87 (\texttt{gpt-4o} with two demonstrations)  & 5 mins for 3,263 GPT prompts & Low \\ \hline
    Embedding & \$0.29 (\texttt{text-embedding-3-large}) 
    & 8 mins for embedding all 12,039 headlines in test set & Moderate \\ \hline
    LoRA & \$3 (Llama-3-8b)  & 2 hours (Llama-3-8b)  & High \\
    \bottomrule
    \end{tabular}
    \caption{Summary of cost, run-time, and effort for different methods. Note that the reported run-time for Prompt-based (GPT-3.5 and GPT-4) and embedding-based methods refer to the time needed to obtain responses or embeddings from OpenAI. Note that the prompt and embedding response time can vary based on the \href{https://platform.openai.com/docs/guides/rate-limits/usage-tiers?context=tier-one}{API usage tiers}, with our results specifically derived from Tier 1. Since we only make prompt inquiries for test data, to maintain a fair comparison, we report the runtime for the embedding method only for headlines in the test data, although we also need to obtain embeddings for headlines in the training data.}
    \label{table:cost_effort}
\end{table}

For the prompt-based method utilizing the OpenAI API, the cost was \$2.16 for \texttt{gpt-3.5-turbo-0125} and \$15.87 for \texttt{gpt-4}, on the test dataset with 3,263 prompts when using two demonstrations. With \texttt{gpt-4} approximately 7.5 times the cost of GPT-3.5. This cost is calculated based on the length (the number of tokens) of the prompt. Runtime for 3,263 prompts is about 5 minutes when using parallelization. There is an inquiry rate/speed limit set by OpenAI that depends on the account tier. Using parallelism could significantly speed up the runtime until the inquiry rate limit is reached.

For embedding-based methods, obtaining text embeddings via the OpenAI API is more economical than using prompt-based methods. At the time of this study (Nov 2024), a \texttt{gpt-4} prompt costs \$2.5 per 1M input tokens and \$10 per 1M output tokens. The embeddings used in our study, \texttt{text-embedding-3-large}, cost only \$0.13 per 1M input tokens, with no output token cost associated with embeddings. The embedding cost of our entire dataset with 77,245 headlines is \$0.29. The reported monetary cost for \texttt{gpt-4} is based on our prompt design, which includes not only the headline but also a description of our problem. Additionally, processing embeddings is much faster (the maximum Tier 1 rate limit---3,000 requests per minute reached) compared to prompt-based methods (500 requests per minute). This efficiency can lead to lower operational costs and easier integration into real-time applications.

For the LoRA fine-tuning method, the GPU requirements depend on factors such as model size, batch size, optimization algorithm, and floating-point precision. In our experiment, full-parameter fine-tuning without LoRA would require over 100 GB of GPU memory. However, LoRA fine-tuning only requires 30 GB of GPU memory. To illustrate the benefit of reduced GPU requirements, we compare two commonly used GPU models for LLM training. Without LoRA, the NVIDIA A100 GPU, which has nearly the largest GPU memory (80 GB), is insufficient for fine-tuning. In contrast, with LoRA, a more affordable NVIDIA V100 GPU with 32 GB of memory is sufficient. The price of an NVIDIA A100 GPU is five times higher than an NVIDIA V100 GPU as of May 2024. Thus, using LoRA significantly reduces costs, making fine-tuning more accessible to researchers and practitioners. In our experiment, each replicate involved approximately 2 hours of fine-tuning the Llama-3-8b model on an NVIDIA A100 GPU with 80 GB memory, costing about \$3 on the cloud computing provider \href{https://www.autodl.com/home}{AutoDL}. We also note that after fine-tuning the model, several cost-efficient techniques are available for deploying it for inference, significantly reducing computational costs and hardware requirements. For instance, \href{https://github.com/ggerganov/llama.cpp}{\texttt{llama.cpp}} allows LLM models to be deployed on machines without a GPU, making implementation more affordable. Moreover, model compression techniques such as quantization, sparsification, and knowledge distillation can drastically cut inference computation costs while maintaining, or only slightly affecting, model performance. For a comprehensive survey on efficient LLM deployment and inference techniques, please refer to \cite{zhou2024survey}.

In summary, we recommend using either embedding-based methods or fine-tuning methods, as they achieve similar performance (at least in our setting), superior to prompt-based methods, and are cost-effective. These methods offer different levels of transparency: prompt-based methods using OpenAI's API are entirely black-box; embedding-based methods use OpenAI's black-box embedding model combined with a user-controlled CTR prediction model; while fine-tuning open-source models provides full transparency. Fine-tuning methods require more significant engineering effort to fine-tune and deploy an LLM than embedding-based methods.

\color{black}

\section{Sensitivity Analysis for Hyperparameter $n^{\text{aux}}$ in LOLA}
\label{appsec:sensitivity}

\begin{figure}[ht]
    \centering
    \includegraphics[width=0.95\linewidth]{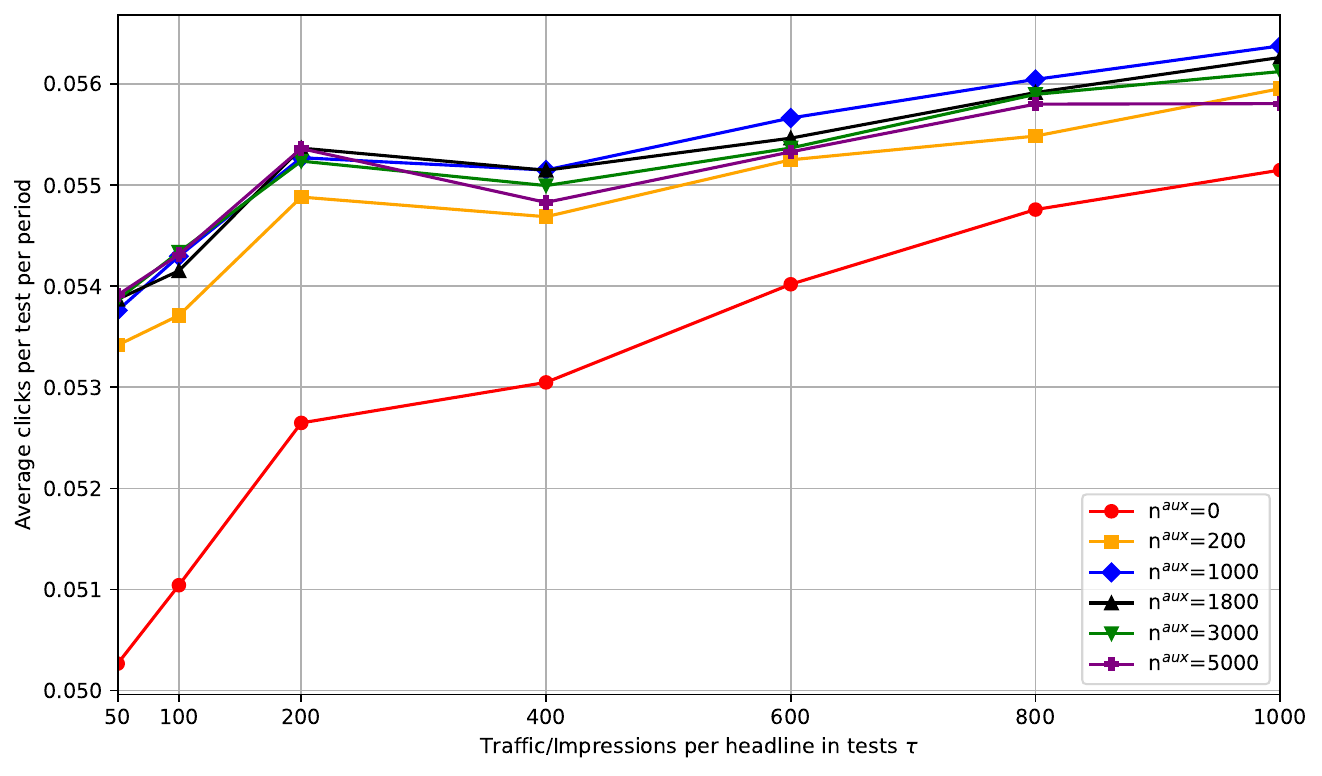}%
    \caption{Result of Sensitivity Analysis for Hyperparameter $n^{\text{aux}}$}
    \label{fig:sensitivity}
\end{figure}

In this part, we document the sensitivity analysis results for the hyperparameter $n^{\text{aux}}$. We begin by fixing the other hyperparameter $\alpha=0.08$, which has been fine-tuned for our setting, and focus solely on $n^{\text{aux}}$. The sensitivity results are visualized in Figure \ref{fig:sensitivity}. To investigate its effect on the final average clicks, we select a wide range of values for $n^{\text{aux}} \in \{0, 200, 1000, 1800, 3000, 5000\}$. Note that this experiment is conducted on the test dataset. When $n^{\text{aux}}=0$, LOLA essentially reduces to the standard UCB, which performs the worst. We observe that LOLA performs best when $n^{\text{aux}}=1000$. This is why we fine-tuned $n^{\text{aux}}$ within the range $\{600, 800, 1000, 1200, 1400\}$ around 1000 in our algorithm. This analysis highlights that the performance of LOLA is robust to the choice of $n^{\text{aux}}$, as $n^{\text{aux}}=1800$ still yields results close to those at $n^{\text{aux}}=1000$.

\section{Details of LOLA Variants and Extensions}
\label{appsec:lola_extensions}
\subsection{LOLA with Best Arm Identification}
\label{appssec:BAI}

In this section, we showcase how the LOLA framework can be used to solve the best arm identification problem. 

The best arm identification problem is a variant of the multi-armed bandit problem, where the goal is to identify the arm (or arms) with the highest reward. This problem is particularly useful in practice when the objective is to identify the best arm as quickly as possible rather than optimizing cumulative reward. Best arm identification has many variants, and our LOLA framework can be effectively integrated into them. We present the general steps of such a best-arm identification algorithm and the integration of LOLA in Algorithm \ref{alg:best_arm_identification}.

\begin{algorithm}[t]
    \caption{LLM-Assisted Best Arm Identification (LLM-BAI)}
    \label{alg:best_arm_identification}
    \begin{algorithmic}[1]
    \State \textbf{LLM Training Phase:} Train a LLM-based prediction model $\mathcal{M}(x)$ for CTRs using historical data samples, where $x$ is the contextual information for arms.
    \State \textbf{Hyperparameter Fine-Tuning:} Use another subset of data to choose the LLM’s equivalent auxiliary sample size, $n^{\text{aux}}$.
    \State \textbf{Online Learning Phase:} Initialize the number of arms $K$, the number of initial pulls $T_k \leftarrow n^{\text{aux}}$, LLM-based CTR prediction $\bar{\mu}_k \leftarrow \mathcal{M}(x_k)$ for all arms $k \in [K]$, and the candidate set as $\mathcal{C} \leftarrow [K]$.
    \While{$|\mathcal{C}| > 1$}
    \State Pull each arm in $\mathcal{C}$ once, denote the observed reward for arm $k$ as $r_k$.
    \State Update the sample average CTRs $\bar{\mu}_k \leftarrow (\bar{\mu}_k T_k + r_k) / (T_k + 1)$, update $T_k \leftarrow T_k + 1$ for all arms $k \in \mathcal{C}$.
    \State Update confidence interval for each arm as $\mathrm{CI}_k \leftarrow [\bar{\mu}_k - C(T_k), \bar{\mu}_k + C(T_k)]$, for all arms $k \in \mathcal{C}$, where $C(T_k)$ is a decreasing and non-negative function of $T_k$.
    \State Eliminate suboptimal arms from $\mathcal{C}$, i.e., eliminate arms from $\mathcal{C}$ if their upper confidence bound is less than the lower confidence bound of some other arms.
    \EndWhile
    \State \textbf{Return:} The set of good arms.
    \end{algorithmic}
\end{algorithm}

We compare the performance of our proposed LLM-BAI to the standard benchmarks, including the standard A/B test, pure BAI algorithm, and pure LLM. Similar to the numerical setting under the regret minimization framework, we treat the reported CTR in the Upworthy dataset as the true expected reward of each headline. Each bandit problem instance corresponds to one news item, and each news headline is viewed as one arm. The return of each arm follows a Bernoulli distribution with the expectation given as the reported CTR. 

We observe that many news items in the Upworthy dataset have headlines with similar CTRs, making the best arm identification problem challenging. In real-world practice, the benefit of choosing the best arm is quite small (or non-existent) when the difference between the best arm and other arms is small (or insignificant). Therefore, we select the $(\mathrm{ST})^2$ algorithm from \cite{mason2020finding} as the pure BAI algorithm, which is tailored to scenarios where the difference between the best arm and the suboptimal arms could be small.\footnote{Note that we can also consider a similar setting in the regret minimization framework, where the planner's goal is to ensure that the regret is below a certain threshold rather than specifying it in comparison to the best arm, as in Equation \eqref{eq:regret}; see \citet{feng2024satisficing} for details.} Specifically, if we denote the real expected return of each arm as $\mu_k$ and the expected return of the best arm as $\mu^* = \max_{k \in [K]} \mu_k$, and given thresholds $\epsilon$ and $\gamma$, and failure rate $\delta$, this algorithm returns a set of arms containing all arms with expected returns better than $\mu^* - \epsilon$, and no arms with expected returns worse than $\mu^* - \epsilon - \gamma$ with high probability. In our case, we set $\epsilon = 0$, $\gamma = 0.005$, and $\delta = 0.2$. It means, with at least $0.8$ probability, the algorithm returns a good set, which does not contain any bad arm with CTR lower than $\mu^* - 0.5\%$. We make slight modifications to the original algorithm for numerical stability; for details, please see our code at \href{https://anonymous.4open.science/r/LOLA_LLM_Assisted_Online_Learning_Algorithm_for_Content_Experiments}{LLM News Github Repository}.

In the implementation, we adopt the same data splitting rule as described in $\S$\ref{ssec:lola_benchmarks}. Specifically, we used 70\% data to LoRA fine-tune the Llama-3-8b CTR prediction model, 10\% data to fine-tune the $n^{\text{aux}}$ parameter, and tested the performance of the LLM-BAI algorithm on the remaining 20\% test dataset. To fine-tune the $n^{\text{aux}}$ parameter, we run the LLM-BAI on the algorithm tuning dataset with different $n^{\text{aux}}$ values in $\{20, 50, 100, 200, 300, 400, 500, 700, 1000\}$, and choose the one with highest success rate in BAI. We finally choose $n^{\text{aux}} = 300$ as the best parameter since it has the highest success rate and the number of pulls is not significantly more than others.

For the natural benchmark pure LLM, we use the same CTR prediction model as LOLA. We also implement the plain BAI without using LLM-based predicted CTR as priors. For the last benchmark, the A/B test method, we allocate a fixed number of pulls to each headline, pulling all headlines an equal number of times. We then calculate the empirical CTR for each headline and select the one with the highest empirical CTR. To evaluate the performance of the A/B test method, we run experiments using several different numbers of pulls per headline, specifically $\{200, 300, 400, 500, 600\}$.

\begin{figure}[ht]
    \centering
    \includegraphics[width=1\linewidth]{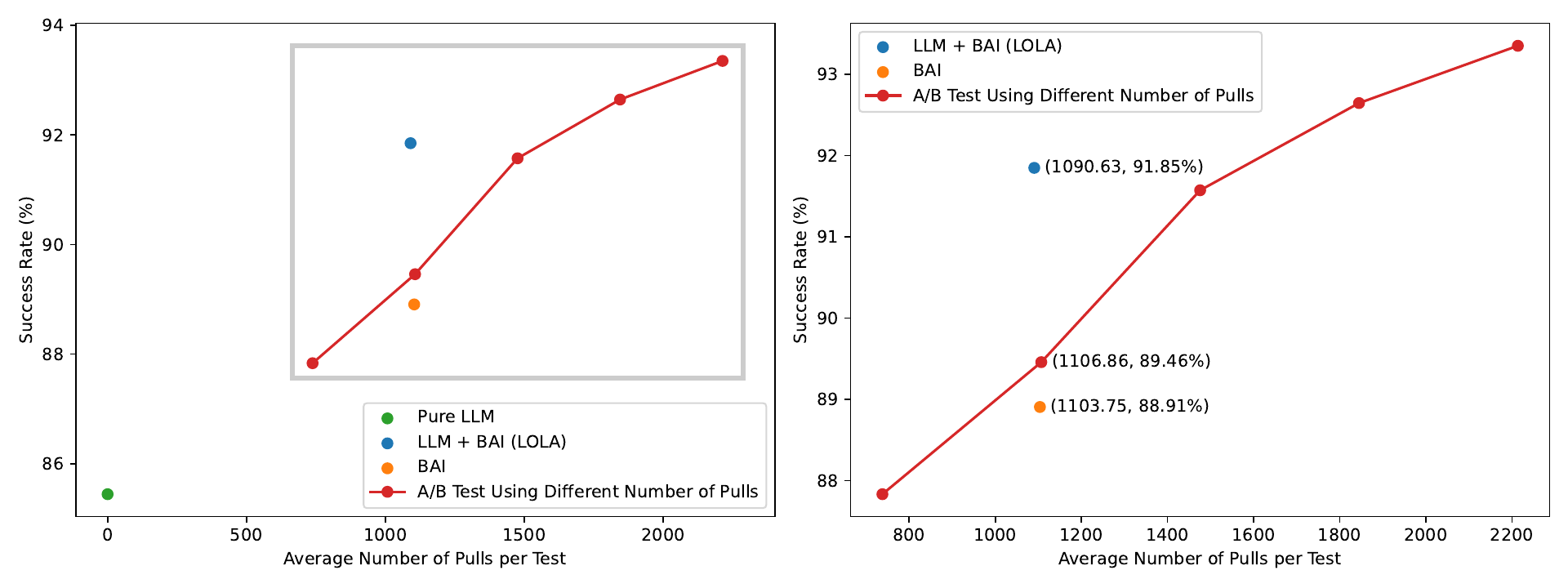}
    \caption{Performance of pure LLM, LOLA, BAI, and A/B test on the test dataset. The X-axis represents the average number of pulls per test (news). The Y-axis represents the success rate. A test is considered successful if all selected headlines (both LOLA and BAI could return multiple good arms) have CTRs above $\mu^* - 0.5\%$. \\The right figure zooms in on the area boxed by the gray rectangle in the left figure to highlight the performance improvement of our LOLA method.}
    \label{fig:BAI_final_performance}
\end{figure}

We report the performance of LLM-BAI and the standard benchmarks in Figure \ref{fig:BAI_final_performance}. Note that the success rate for pure LLM here is higher than the accuracy (46.86\%) reported in Figure \ref{fig:lora_result}, as we use a more lenient definition of success. Our results indicate that our LOLA algorithm outperforms both the pure BAI method and the A/B test method in terms of success rate when using a similar number of pulls per test. Specifically, LOLA improves the success rate by an absolute 2.39\% and saves 1.47\% of pulls compared to the A/B test. Compared to the plain BAI method, LOLA improves the success rate by an absolute 2.94\% and saves 1.19\% of pulls.

\subsection{LOLA with Thompson Sampling}
\label{appsec:ts}
We present a version of LOLA (LLM-TS) that combines LLM and Thompson Sampling and compare its numerical performance with standard Thompson Sampling and LLM-2UCBs. 

LLM-TS extends the standard TS algorithm by initializing an LLM-based prior. See Algorithm \ref{alg:llm-ts} for details. Similar to LLM-2UCBs, here also we need to fine-tune the hyperparameter, $n^{\text aux}_k$ in this case, which is the LLM's equivalent auxiliary sample size. As in the main analysis, we fine-tune this parameter on the fine-tuning dataset and choose from the set  In our numerical experiment, we consider the  $n^{\text aux}_k \in \{600, 800, 1000, 1200, 1400\}$ and find that $n^{\text aux}_k =1200$ generates the highest clicks in the algorithm tuning dataset (which we then use in all our numerical experiments). Notice that this number is quite close to 1000, which was the number used in LLM-2UCBs. There is no significant difference using either $1000$ or $1200$ in clicks on the tuning dataset. The Beta prior $(\alpha^0_k,\beta^0_k)$ implies a CTR mean of $\alpha^0_k/(\alpha^0_k+\beta^0_k)$ and variance of $\alpha^0_k\beta^0_k/((\alpha^0_k+\beta^0_k)^2(\alpha^0_k+\beta^0_k+1))$. We use the mean and variance from the training data to initialize the priors to $\alpha^0_k=1.38$ and $\beta^0_k=96.11$ for all $k\in[K]$ in our experiment.

\begin{algorithm}[t]
\caption{LLM-Assisted Thompson Sampling (LLM-TS)}
\label{alg:llm-ts}
\begin{algorithmic}[1]
\State \textbf{LLM Training Phase:} Train a LLM-based prediction model $\mathcal{M}(x)$ for CTRs using historical data samples, where $x$ is the contextual information for arms.
\State \textbf{Hyperparameter Fine-Tuning:} Use another subset of data to select the LLM's equivalent auxiliary sample size, $n^{\text aux}$ and the standard TS prior $(\alpha^0,\beta^0)$.
\State \textbf{Online Learning Phase:}  Initialize the number of periods $T$, number of arms $K$, LLM-based CTR prediciton $\bar{\mu}^{\text aux}_k \leftarrow \mathcal{M}(x_k)$, the parameters in Beta priors $(\alpha^1_k,\beta^1_k)\leftarrow (\alpha^0,\beta^0) + (n^{\text aux}\bar{\mu}^{\text aux}_k, n^{\text aux}(1-\bar{\mu}^{\text aux}_k))$ for all arms $k\in[K]$.

\For{$t = 1$ \textbf{to} $T$}
    \State Sample $\theta_k^t \sim \text{beta}(\alpha^t_k,\beta^t_k)$ for all arms $k\in[K]$.
    \State Play the arm $a_t = \arg\max_{k\in[K]} \theta_k^t$.
    \State Observe the payoff $r_{a_t}$.
    \State Update $(\alpha^{t+1}_k,\beta^{t+1}_k) \leftarrow (\alpha^t_k,\beta^t_k) + (r_{a_t}, 1-r_{a_t})$ if $a_t=k$; update $(\alpha^{t+1}_k,\beta^{t+1}_k) \leftarrow (\alpha^t_k,\beta^t_k) $ if $a_t\neq k$.
\EndFor
\end{algorithmic}
\end{algorithm}

\begin{figure}[ht]
    \centering
    \includegraphics[width=0.95\linewidth]{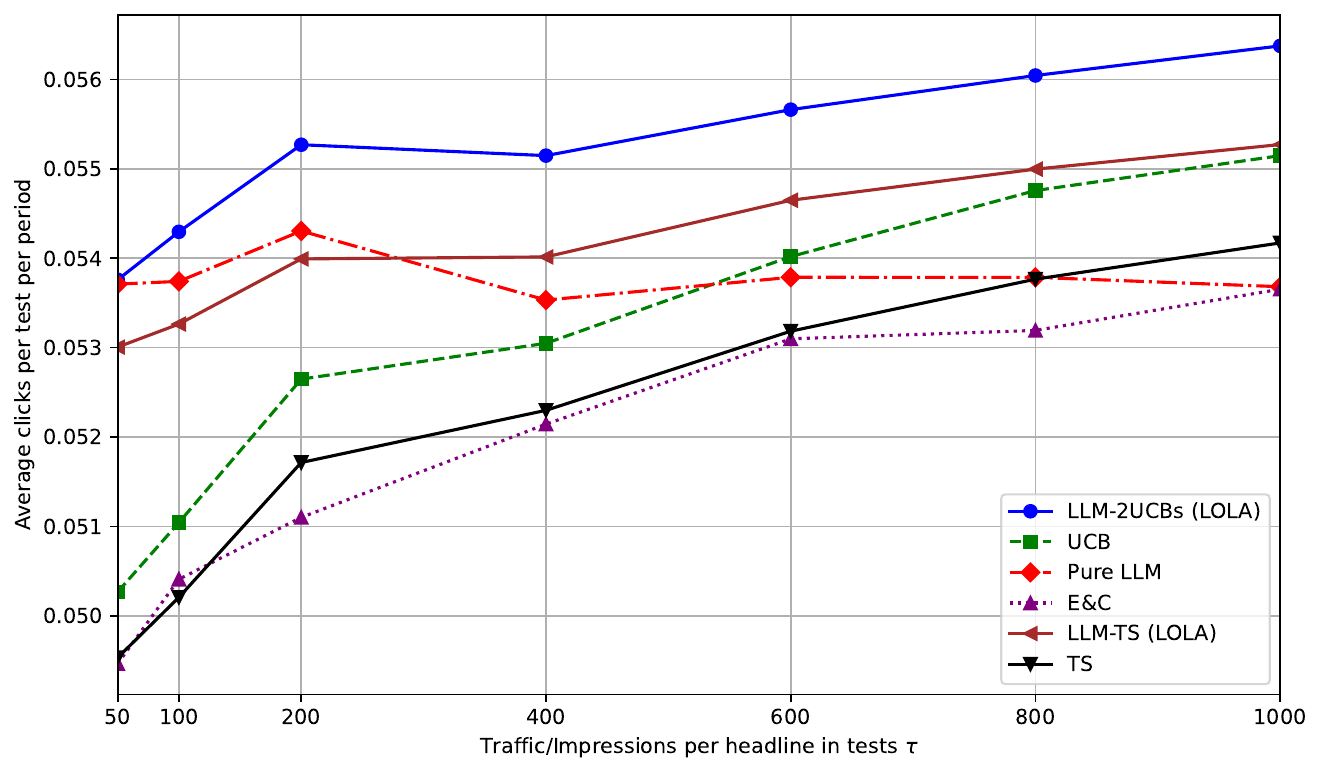}
    \caption{Average clicks per experiment per period under different time horizon multipliers. Note that the Y-axis captures the average clicks per test per period. For instance, if there is a test with two headlines receiving 1 and 2 clicks, conducted under $x=100$, then the average click per period in this test is calculated as $(1+2)/100=0.03$. The Y value is simply the average of this number $0.03$ over all tests. This measure scales well with the platform's total clicks in tests because headlines in different tests with different numbers of headlines take the same weight in this measure. }
    \label{fig:bandit_regret_ts}
\end{figure}

We now present the results on the performance of both LLM-TS and the standard TS algorithm on the test dataset in Figure \ref{fig:bandit_regret_ts}. For comparison, we also show the performance of LLM-2UCBs and the other benchmarks. We also document the pairwise comparison of different algorithms and report the relative percentage reward improvement and the p-value from the t-tests in Table \ref{tab:bandit_regret_ts}. We find that UCB-based algorithms perform better than TS-based algorithms in both the LLM-assisted version and the standard version. This observation is not uncommon because it is widely known that TS suffers from over-exploration \citep{min2020policy}; in other words, UCB's upper confidence shrinks faster, leading it to explore more effectively). As a result, while it is possible to implement a Bayesian bandit such as an LLM-Assisted Thompson Sampling within our LOLA framework, at least in our setting, UCB-based algorithms perform better.


\begin{sidewaystable}[ht] 
\centering
\begin{tabular}{
    >{\raggedleft\arraybackslash}c
    |S[table-format=1.2]
    S[table-format=1.2]
    S[table-format=1.2]
    S[table-format=1.2]
    S[table-format=1.2]
    S[table-format=1.2]
    S[table-format=1.2]
    S[table-format=1.2]
}
\toprule
{Parameter $\tau$} & 
\multicolumn{1}{>{\raggedleft\arraybackslash}p{2cm}}{\scriptsize LLM-2UCBs vs. UCB} & 
\multicolumn{1}{>{\raggedleft\arraybackslash}p{2cm}}{\scriptsize LLM-2UCBs vs. Pure LLM} & 
\multicolumn{1}{>{\raggedleft\arraybackslash}p{2cm}}{\scriptsize LLM-2UCBs vs. E\&C} & 
\multicolumn{1}{>{\raggedleft\arraybackslash}p{2cm}}{\scriptsize LLM-2UCBs vs. LLM-TS} & 
\multicolumn{1}{>{\raggedleft\arraybackslash}p{2cm}}{\scriptsize LLM-2UCBs vs. TS} & 
\multicolumn{1}{>{\raggedleft\arraybackslash}p{2cm}}{\scriptsize UCB vs. Pure LLM} & 
\multicolumn{1}{>{\raggedleft\arraybackslash}p{2cm}}{\scriptsize UCB vs. E\&C} &
\multicolumn{1}{>{\raggedleft\arraybackslash}p{2cm}}{\scriptsize UCB vs. LLM-TS} \\
\midrule
50 & 6.95$^{****}$ & 0.09 & 8.69$^{****}$ & 1.42$^{**}$ & 8.54$^{****}$ & -6.41$^{****}$ & 1.62 & -5.17$^{****}$ \\
100 &  6.38$^{****}$ & 1.03$^{**}$ & 7.72$^{****}$ & 1.94$^{****}$ & 8.15$^{****}$ & -5.02$^{****}$ & 1.26 & -4.17$^{****}$ \\
200 & 4.98$^{****}$ & 1.78$^{****}$ & 8.16$^{****}$ & 2.37$^{****}$ & 6.88$^{****}$ & -3.05$^{****}$ & 3.03$^{**}$ & -2.49$^{****}$\\
400 & 3.96$^{****}$ & 3.02$^{****}$ & 5.76$^{****}$ & 2.10$^{****}$ & 5.45$^{****}$ & -0.90 & 1.74$^{*}$ & -1.79$^{****}$ \\
600 & 3.04$^{****}$ & 3.49$^{****}$ & 4.83$^{****}$ & 1.86$^{****}$ & 4.66$^{****}$ & 0.43 & 1.73$^{**}$ & -1.15$^{***}$ \\
800 & 2.35$^{****}$ & 4.20$^{****}$ & 5.36$^{****}$ & 1.91$^{****}$ & 4.24$^{****}$ & 1.81$^{***}$ & 2.94$^{****}$ & -0.44  \\
1000 & 2.23$^{****}$ & 5.02$^{****}$ & 5.08$^{****}$ & 2.00$^{****}$ & 4.07$^{****}$ & 2.73$^{****}$ & 2.79$^{****}$ & -0.23 \\
\end{tabular}

\vspace{1cm}  

\begin{tabular}{
    >{\raggedleft\arraybackslash}c
    |S[table-format=1.2]
    S[table-format=1.2]
    S[table-format=1.2]
    S[table-format=1.2]
    S[table-format=1.2]
    S[table-format=1.2]
    S[table-format=1.2]
    S[table-format=1.2]
}
\toprule
{Parameter $\tau$} & 
\multicolumn{1}{>{\raggedleft\arraybackslash}p{2cm}}{\scriptsize UCB vs. TS} & 
\multicolumn{1}{>{\raggedleft\arraybackslash}p{2cm}}{\scriptsize Pure LLM vs. E\&C} & 
\multicolumn{1}{>{\raggedleft\arraybackslash}p{2cm}}{\scriptsize Pure LLM vs. LLM-TS} & 
\multicolumn{1}{>{\raggedleft\arraybackslash}p{2cm}}{\scriptsize Pure LLM vs. TS} & 
\multicolumn{1}{>{\raggedleft\arraybackslash}p{2cm}}{\scriptsize E\&C vs. LLM-TS} & 
\multicolumn{1}{>{\raggedleft\arraybackslash}p{2cm}}{\scriptsize E\&C vs. TS} & 
\multicolumn{1}{>{\raggedleft\arraybackslash}p{2cm}}{\scriptsize LLM-TS vs. TS} \\
\midrule
50 & 1.48$^{*}$ & 8.59$^{****}$ & 1.33$^{**}$ & 8.44$^{****}$ & -6.68$^{****}$ & -0.14 & 7.02$^{****}$ \\
100 & 1.67$^{***}$ & 6.61$^{****}$ & 0.90$^{**}$ & 7.04$^{****}$ & -5.36$^{****}$ & 0.40 & 6.09$^{****}$  \\
200 & 1.80$^{****}$ & 6.27$^{****}$ & 0.58 & 5.01$^{****}$ & -5.36$^{****}$ & -1.19 & 4.41$^{****}$ \\
400 & 1.43$^{****}$ & 2.66$^{***}$ & -0.89$^{**}$ & 2.36$^{****}$ & -3.47$^{****}$ & -0.30 & 3.28$^{****}$  \\
600 & 1.57$^{****}$ & 1.30 & -1.58$^{****}$ & 1.13$^{*}$ & -2.84$^{****}$ & -0.16 & 2.76$^{****}$ \\
800 & 1.84$^{****}$ & 1.12 & -2.20$^{****}$ & 0.04 & -3.28$^{****}$ & -1.07$^{*}$ & 2.29$^{****}$  \\
1000 & 1.80$^{****}$ & 0.05 & -2.88$^{****}$ & -0.90 & -2.93$^{****}$ & -0.96$^{*}$ & 2.03$^{****}$  \\
\end{tabular}
\caption{Percentage \% reward improvement between two algorithms with significance levels: $^{*}$ $p\leq$ 0.05, $^{**}$ $p\leq$ 0.01, $^{***}$ $p\leq$ 0.001, $^{****}$ $p\leq$ 0.0001. For example, under the impression per headline equal to 50, the average click per test of LOLA and UCB is 0.053760 and 0.050267, respectively; so the relative improvement is calculated as ${0.053760}/{0.050267} - 1 = 6.95\%$. Parameter $\tau$ represents the average impression per headline to scale the time horizon $T=\tau\times K$.}
\label{tab:bandit_regret_ts}
\end{sidewaystable}

\subsection{LOLA with Contextual Information}
\label{appsec:algs}

We now introduce two additional LOLA algorithms that leverage contextual (i.e., user-specific features) information.

For linear contextual bandits, we assume that at each round, we observe the context vector $x_{t,k} \in \mathbb{R}^d$ for each arm $k \in [K]$. This context information can include LLM embedding vectors and user-specific features, among other features. The random reward of arm $k$ at round $t$ is denoted by $r_{t,k}$. After pulling one arm $a_t$ at each round, only the reward of that arm, $r_{a_t}$, is revealed. We make the linear realizability assumption that the expected reward is linear in the context vectors, which means $\bE[r_{t,k} | x_{t,k}] = x_{t,k}^\top \theta^*$. The weight vector $\theta^* \in \mathbb{R}^d$ is unknown and needs to be learned. Additionally, we assume access to a historical dataset of size $M$, denoted by $\{(x_{t,k}, r_t)\}_{t=-M}^{-1}$. With these definitions, we now present our proposed Algorithm \ref{alg:llm-linUCB}, named LLM-Assisted Linear Contextual Bandit. The LLM-LinUCB algorithm extends LinUCB \cite{chu2011contextual}. Under the linear realizability assumption, this algorithm achieves asymptotically optimal regret, matching the lower bound up to logarithmic factors. Note that this algorithm does not have an auxiliary sample hyperparameter $n^{\text aux}$ because the historical sample size $M$ can automatically adjust the confidence bound via Gram matrix $A$, assuming the same linear reward function during the online learning phase.

\begin{algorithm}[ht]
\caption{LLM-Assisted Linear Contextual Bandit (LLM-LinUCB)}
\label{alg:llm-linUCB}
\begin{algorithmic}[1]
\State \textbf{Embedding-based Classifier Training Phase:} Calculate $A \leftarrow I_d + \sum_{t=-M}^{-1}x_{t,a_t} x_{t,a_t}^\top$, where $I_d$ is the $d$-by-$d$ identity matrix to avoid singular $A$, and $b \leftarrow  \sum_{t=-M}^{-1}x_{t,a_t} r_t$. 
\State \textbf{Hyperparameter Fine Tune:}  $\alpha \in \mathbb{R}^+$ for controlling the upper bound. 
\State \textbf{Online Learning Phase:} Initialize the number of periods $T$, number of arms $K$.
\For{$t = 1$ \textbf{to} $T$}
    \State Get weight estimator: $\theta_t \leftarrow A^{-1}b$
    \State Observe $K$ context vectors, $x_{t,1}, x_{t,2}, \dots, x_{t,K} \in \mathbb{R}^d$, including text embeddings, user-specific feature.
    \For{$k = 1, 2, \dots, K$}
        \State $\hat{r}_{t,k} \leftarrow \theta_t^\top x_{t,k} + \alpha \sqrt{x_{t,k}^\top A^{-1} x_{t,k}}$ (Computes the upper confidence bound)
    \EndFor
    \State Choose action $a_t = \arg\max_{k\in[K]} \hat{r}_{t,k}$.
    \State Observe reward $r_{a_t}$.
    \State $A \leftarrow A + x_{t,a_t} x_{t,a_t}^\top$.
    \State $b \leftarrow b + x_{t,a_t} r_{a_t}$.
\EndFor
\end{algorithmic}
\end{algorithm}

The algorithm proceeds as follows: During the classifier training phase, leveraging historical data, we prepare the matrix $A$ and vector $b$ to estimate the linear weight vector. We need an LLM-based embedding model to extract embedding vectors from the original text information. The hyperparameter fine-tuning is similar to LLM-2UCBs but with only the upper bound control parameter $\alpha$. Specifically, one can use the bisection search algorithm to find the optimal $\alpha$ that maximizes the total reward on this fine-tuning dataset. The online learning phase remains the same as the standard LinUCB.

The advantages of LLM-LinUCB are its transparency due to the linearity assumption and ease of execution with only one hyperparameter, as well as the theoretical soundness of its optimal regret results. However, there are two issues with the linearity assumption. First, although we have shown that linear functions almost fully leverage text embeddings from OpenAI to achieve high accuracy (there is almost no difference compared to MLP), the embedding itself sacrifices a lot of information, as demonstrated by the significant outperformance of LoRA fine-tuning over the embedding-based approach. Second, once concatenated with user contexts, it is unclear whether the linear function is a good approximation of the rewards model (and this requires further numerical study). In addition, LLM-LinUCB cannot accommodate LoRA fine-tuned open-source LLM, which is a complex nonlinear reward model. 

To alleviate the limitations of the linear reward assumption and allow general LoRA fine-tuned open-source LLM with user-specific features, we introduce another general LLM-assisted contextual bandit algorithm (Algorithm \ref{alg:llm-falcon}), which extends the FAst Least-squares-regression-oracle CONtextual bandits (FALCON) algorithm proposed by \cite{simchi2022bypassing}. This contextual bandit algorithm generalizes the linear realizability assumption to a more flexible framework, assuming that the true prediction model $f^* \in \mathcal{F}$. The functional class $\mathcal{F}$ can be very general and is not limited to linear models. This assumption is reasonable and promising because LLMs have demonstrated their capability across a wide range of tasks. Essentially, $\mathcal{F}$ can be viewed as a class of LLM models, and the goal is to find the true LLM model $f^*$ that performs best for our task, i.e., $f^*(x_{t,k},k) = \mathbb{E}[r_{t,k} | x_{t,k}]$ for any $k \in [K]$ and $x_{t,k} \in \mathcal{X}$. Under this setting, the context information is not restricted to real numbers and can include more general data, such as text information. We now present the LLM-FALCON algorithm.

\begin{algorithm}[ht]
\caption{LLM-Assisted FAst Least-squares-regression-oracle CONtextual bandits (LLM-FALCON)}
\label{alg:llm-falcon}
\begin{algorithmic}[1]
\State \textbf{LLM Training Phase:} Obtain the historical data $\{(x_{t,k}, r_t)\}_{t=-M}^{0}$, and construct an LLM $\mathcal{F}$ class for training.
\State \textbf{Hyperparameter Fine Tuning:} Confidence control parameter $\delta$.
\State \textbf{Online Learning Phase:} Initialize $T$, $K$, and the epoch schedule $0 = \tau_0 < \tau_1 < \tau_2 < \cdots$.
\For{epoch $m = 1, 2, \dots$}
    \State Set $\gamma_m = (1/30) \sqrt{(K \tau_{m-1}) / \log(|\mathcal{F}| \tau_{m-1} / \delta)}$ (for epoch 1, $\gamma_1 = 1$).
    \State Compute $\hat{f}_m = \arg\min_{f \in \mathcal{F}} \sum_{t=-M}^{\tau_{m-1}} (f(x_{t,a_t}, a_t) - r_t)^2$ via the offline least squares oracle.
    \For{round $t = \tau_{m-1} + 1, \dots, \tau_m$}
        \State Observe contexts $x_{t,k} \in \mathcal{X}$,$\forall k\in [K]$.
        \State Compute $\hat{f}_m(x_{t,k}, k)$ for each arm $k \in [K]$. Let $\hat{a}_t = \arg\max_{k \in [K]} \hat{f}_m(x_{t,k}, k)$.
        \State Define
        \[
        p_t(k)= \begin{cases} 
        \frac{1}{K + \gamma_m (\hat{f}_m(x_{t,\hat{a}_t}, \hat{a}_t) - \hat{f}_m(x_{t,k}, k))}, & \forall k \neq \hat{a}_t, \\
        1 - \sum_{k \neq \hat{a}_t} p_t(k), & k = \hat{a}_t.
        \end{cases}
        \]
        \State Sample $a_t \sim p_t(\cdot)$ and observe reward $r_{a_t}$.
    \EndFor
\EndFor
\end{algorithmic}
\end{algorithm}

At the beginning of each epoch, we retrain the LLM using all available data up to that point, as detailed in Line 6. For simplicity, this algorithm assumes no optimization error during training, utilizing the offline least squares oracle instead. However, in practice, achieving zero optimization error is impossible, and one can incorporate this error into the final regret term using the regret triangle decomposition. Although this algorithm can handle original text information and fits into a general LLM model (not restricted to linear models), the primary challenge is the cost of training the LLM at each epoch because training a full LLM is typically time-consuming and costly. One potential solution, as investigated in our paper, is to fine-tune the LLM using LoRA instead of fully retraining it with new data. However, the empirical performance of this method remains unclear.

After getting an LLM trained on new data, then the key step in this algorithm is using the ``inverse proportional to the gap" rule to balance exploration and exploitation. This rule is not new \citep{foster2020beyond,simchi2022bypassing} and can effectively reduce a contextual bandit problem to a regression problem. At each round, it first identifies the empirically best arm $\hat{a}_t$ and then computes the gap in expected reward between this arm and the others. The sampling probability $p_t(k)$ is inversely proportional to this gap, meaning that a larger predicted reward $\hat{f}_m(x_{t,k}, k)$ results in a higher probability of being pulled/exploited. The gradually increasing parameter $\gamma_m$ (as a sequential exploration and exploitation balancer) guarantees the algorithm to explore more in the beginning rounds and exploits more in later rounds.

\end{appendices}


\end{bibunit}
\end{document}